\documentclass[aps,twocolumn]{revtex4-1}

\usepackage{graphicx}
\usepackage{xcolor}
\usepackage[english]{babel}
\usepackage{amsmath}
\usepackage{amssymb}
\usepackage{bm}
\usepackage{makecell}
\usepackage{array}
\usepackage[normalem]{ulem}

\definecolor{darkgreen}{RGB}{0, 150, 0}

\def\1{{\mathds{1}}}

\begin{document}

\author{Alice E. A. Allen}
\affiliation
{Department of Physics and Materials Science, University of Luxembourg, L-1511 Luxembourg City, Luxembourg}
\affiliation
{Center for Nonlinear Studies, Los Alamos National Laboratory, Los Alamos, NM, USA}
\affiliation
{Theoretical Division, Los Alamos National Laboratory, Los Alamos, NM, USA}

\author{Alexandre Tkatchenko}
\email{alexandre.tkatchenko@uni.lu}
\affiliation
{Department of Physics and Materials Science, University of Luxembourg, L-1511 Luxembourg City, Luxembourg}


\title[]{Constructing Effective Machine Learning Models for the Sciences: \\A Multidisciplinary Perspective}

\begin{abstract}
Learning from data has led to substantial advances in a multitude of disciplines, including text and multimedia search, speech recognition, and autonomous-vehicle navigation. Can machine learning enable similar leaps in the natural and social sciences? This is certainly the expectation in many scientific fields and recent years have seen a plethora of applications of non-linear models to a wide range of datasets. However, flexible non-linear solutions will not always improve upon manually adding transforms and interactions between variables to linear regression models.  We discuss how to recognize this before constructing a data-driven model and how such analysis can help us move to intrinsically interpretable regression models. Furthermore, for a variety of applications in the natural and social sciences we demonstrate why improvements may be seen with more complex regression models and why they may not.  
\end{abstract}

\maketitle

\section{Introduction}

Machine learning (ML) has transformed sciences and technologies by offering excellent predictive performance for a range of applications~\cite{Silver2018, Noe2019,Senior2020,Gomez2018,Faber2016,Lampa2014, Keith2021, Deringer2019-ss, Noe2020, Athey2019, Goodall2020, Greener2022, Widmann2022,Ramsundar2019,Johri2020,Rupp2012,Duvenaud2015,Sanchez2018,Altae2017}. However, complex ``black box'' solutions, like neural networks (NNs), are not always the most appropriate models~\cite{Schulz2020, Ghiringhelli2015, Rudin2019, Allen2022, Volovici2022,Obermeyer2019,Szegedy2014,Mehrabi2021}. We outline the circumstances when improvements in predictive performance are likely to occur with non-linear solutions and when intrinsically interpretable linear models are the most suitable choice. In doing so, we aim to help researchers evaluate the likely outcome of using a regression model with a more flexible functional form, and  understand when improvements may be seen.  

 Whilst the contents of the datasets used in different fields vary, the methods applied for data analysis overlap significantly and machine learning has become a unifying factor between disciplines. For example, text mining has been used in a wide variety of applications from predicting structures in material science to extracting information in legislative bills~\cite{Ceder2006,Fluck2014,Wilkerson2017}. Convolutional NNs inspired by image-recognition models have been used to accurately predict static and dynamic properties of molecules~\cite{Schutt2017}. Transformer models have examined political texts \cite{Widmann2022}, predicted DNA enhancers from sequence information \cite{Le2021} and been used for molecular and material property prediction \cite{Wang2021,Unke2021}. Despite the similarities in techniques, discussions often remain isolated to individual subject areas. We look to overcome this bias by using real and simulated datasets from across the natural and social sciences (see Fig. 1). A key difference between the use of modelling in these fields is that the focus in the social sciences is primarily to provide a causal explanation~\cite{Athey2019, Shmueli2010}. In the natural sciences, whilst some modelling is carried out for explanatory purposes, predictive performance is often the key metric of success. By bringing together discussions and examples from a range of fields, we seek to provide a more diverse viewpoint regarding the required complexity for a predictive model. 

\begin{figure*}[htb]
    \centering
    \includegraphics[width=7.00in]{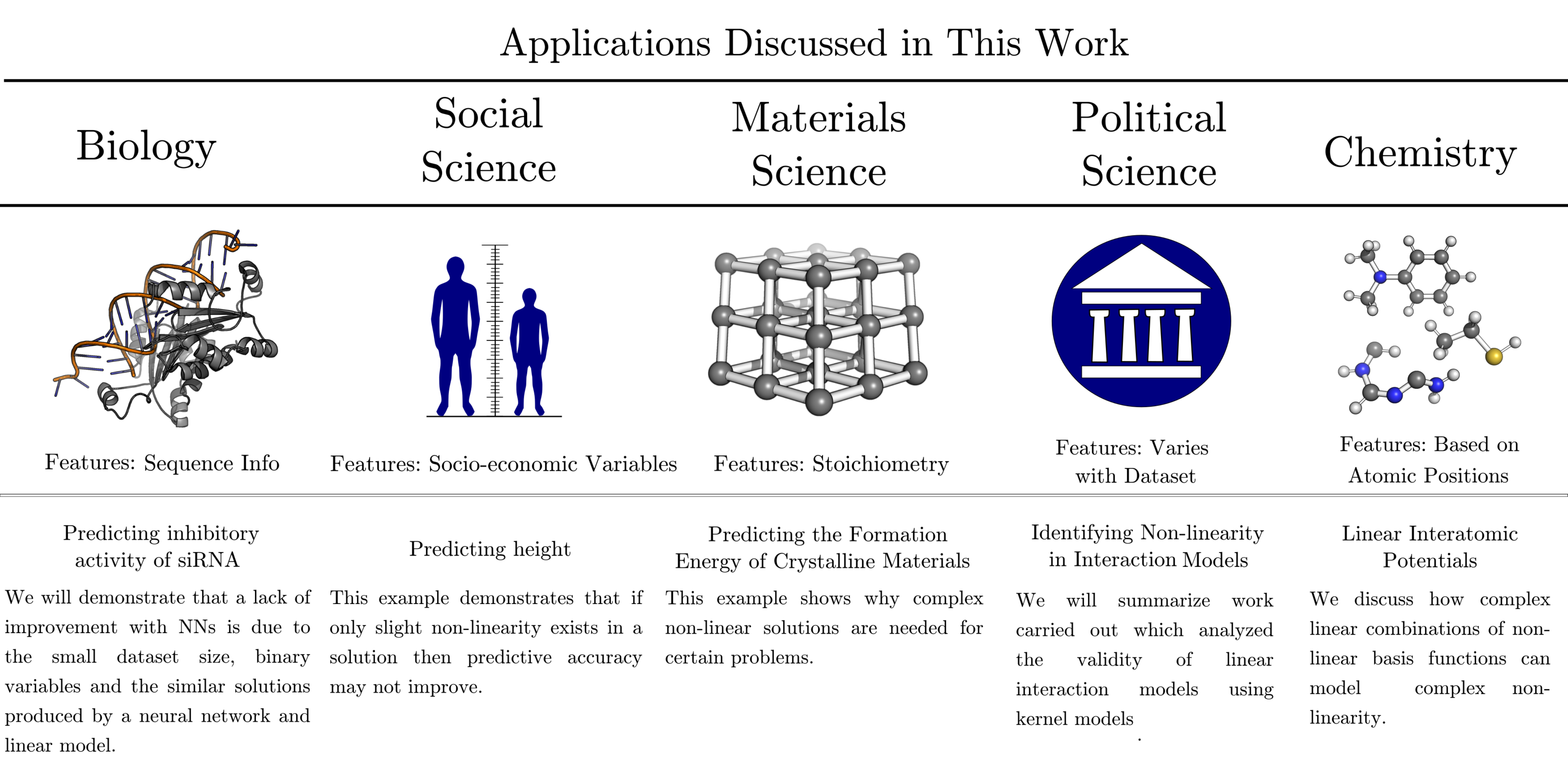}
    \caption{We cover a variety of applications and discussions from both the natural and social sciences in this work. The figure shows this and summarizes the variables used in the predictive models created. By drawing on examples from a range of disciplines, we highlight the similar characteristics seen in many datasets.  
}
    \label{fig:Summary_diag}
\end{figure*}

 A fair comparison between linear and non-linear models requires discussing further differentiating factors. Linear regression can be used with an extremely large number of basis functions, or with kernel transforms, to recreate highly complex non-linearity~\cite{Hogg_2021}. In this discussion, we are focusing on linear combinations of non-linear basis functions which are simplistic enough to be ``intrinsically interpretable''. These kind of linear approximations are traditionally used as statistical models across the sciences to explain relationships~\cite{Shmueli2010}. Finding interpretable linear models relies on the existence of appropriate features whose contributions can be added up to achieve high levels of predictive accuracy. Whilst it is not possible to strictly define interpretable linear models, they will have the following characteristics. Firstly, there will be a small (on the order of tens or hundreds) of variables, there will be a limited set of transforms and the variables used will have clear meaning. Using linear models with thousands of basis functions to describe complex non-linearity is discussed in Section \ref{sec:chem}. We emphasize that whether non-linear or linear models are employed, fundamental statistical learning techniques, such as cross-validation and regularization, should always be used to ensure accurate models are produced that can generalize to unseen data.

The flexible functional forms of non-linear ML methods can result in the dramatic enhancement of predictive  accuracy~\cite{Schutt2017,Deringer2019-ss,Athey2019,Jha2018,Batzner2022}. This is particularly true in cases where there is plentiful data and domain knowledge can be integrated into models~\cite{Silver2018,Senior2020,Chmiela:2017ff,Chmiela2018}. However, there are multiple different factors that may impact if a non-linear model will improve over an interpretable linear model. For example, the size, processing and nature of the variables used in a dataset can reduce the possible improvements in accuracy. Alternatively, the problem itself may be sufficiently well approximated by a simple linear model, particularly with the addition of interaction terms and transforms, that moving to a flexible functional form does not help~\cite{Layard2014, Brambor2006,Allen2022,Blankertz2007}. It may also be that causal explanations or interpretability are the primary goal of a project. Although interpretability with non-linear models is possible, if a simple linear model can instead be used this may be preferable~\cite{Muller2003}. We will discuss these factors, along with the circumstances when non-linear ML methods will significantly improve results. Understanding these characteristics of a problem is important, as it can prevent researchers unnecessarily searching for overly complex non-linear solutions and choosing simplicity where possible~\cite{Muller2003}.

Whilst in theory data of any complexity can be described with a linear model provided that a sufficient number of appropriate features are used, in practice finding a manageable number of these features for many applications is impossible. One perspective on NNs is that they automatically find the optimal non-linear features before finally combining them in a linear layer. Applications which contain complex non-linearity that can only be described using deep learning models include image analysis tools, language models and scientific models such as AlphaFold, SchNet, PauliNet, and Boltzmann generators~\cite{Senior2020,Schutt2017,Hermann2020,Noe2019}. For these applications, deep learning models offer an obvious advantage over intrinsically interpretable methods. Post-hoc interpretability methods then become an essential tool for improving understanding~\cite{Samek2021}. Techniques, such as LIME (Local Interpretable Model-agnostic Explanations), can break down predictions into variable contributions to describe the reasons for an individual result~\cite{Strumbelj2013, Ribeiro2016, Lundberg2017}. Alternatively techniques like SHAP (SHapley Additive exPlanations) or partial dependence can visualize the non-linearity and interactions that exist between variables ~\cite{Friedman2001, Goldstein2015,Apley2020, Hooker2007, Lundberg2017, Samek2021}. For NNs, there is also an ever-expanding set of interpretability methods, such as layer-wise relevance propagation~\cite{LRP} and spectral relevance analysis~\cite{CleverHans,Wilming2022}. A detailed discussion regarding interpretability of deep NNs, and the variety of methods available is presented in Ref.~\citenum{Samek2021}.
 
\begin{figure*}[htb!]
    \centering
    \includegraphics[width=7.0in]{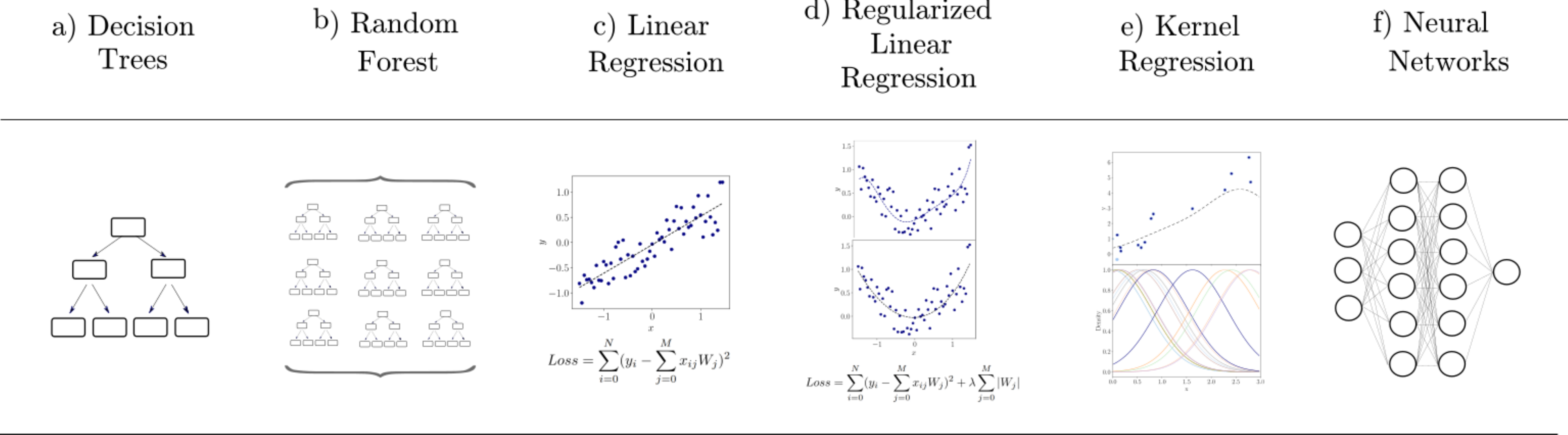}
    \caption{Examples of popular ML models that can be used for regression problems: a) decision trees, which use simple decision rules to provide predictions, b) random forest which uses an ensemble of decision trees, c) linear regression which uses a linear approach for modelling, d) linear regression with regularization which can prevent overfitting occurring, e) kernel methods which map the variables into a \emph{feature} space where the regression problem becomes linear and f) NN models that use a network of functions to form predictions.  }
    \label{fig:ML_reg_model}
\end{figure*}

With the growth in ML interpretability methods, it is now clear that interpretability and non-linear methods are not incompatible~\cite{Samek2021}. In fact, there are many cases where  linear models become difficult to interpret. When the number of basis functions gets very large, or the features used are subject to an extensive level of transforms, linear models are no longer `intrinsically' interpretable models. Additionally, even simple linear models can be misinterpreted~\cite{Brambor2006, Hainmueller2019, Athey2019, Wilming2022}. Furthermore, the coefficients of a linear model cannot be analysed without a measure of the coefficient's error as there may be a large space of possible solutions available. When regularization is added to a linear model, interpretation becomes even more complex and casual conclusions can then be particularly problematic~\cite{Athey2019}. This is especially true when variables are  correlated~\cite{Wilming2022}. In these cases, interpreting a linear model is not necessarily a simple task. However, simplistic linear models, with a small number of features and transforms, have been the focus for explanatory modelling in many scientific disciplines. Other intrinsically interpretable non-linear models exist, but the compact form of an interpretable and predictive linear model remains appealing.

\section{When Linear Models Are All You Need}
Non-linear ML models have a flexible functional form which can model complex relationships between independent variables. However, there are a number of reasons why this does not always lead to an improvement in predictive performance over linear combinations of non-linear basis functions (which we refer to as linear models throughout this work). 
 
There are many different types of models that can be used for predicting the properties of a dataset, see Fig.~\ref{fig:ML_reg_model}. The discussion that follows is not restricted to a specific type of non-linear model and is generally applicable to many existing data-driven methods. A visual representation of the types of regression models available is shown in Fig.~\ref{fig:ML_reg_model}.

It is also important to note that this work focuses on predictive models. Other types of ML tasks, such as generative and reinforcement learning, generally require more autonomy to be given to the model and flexible functional forms are therefore more likely to be needed.

\begin{figure*}[htb]
    \centering
    \includegraphics[width=7.0in]{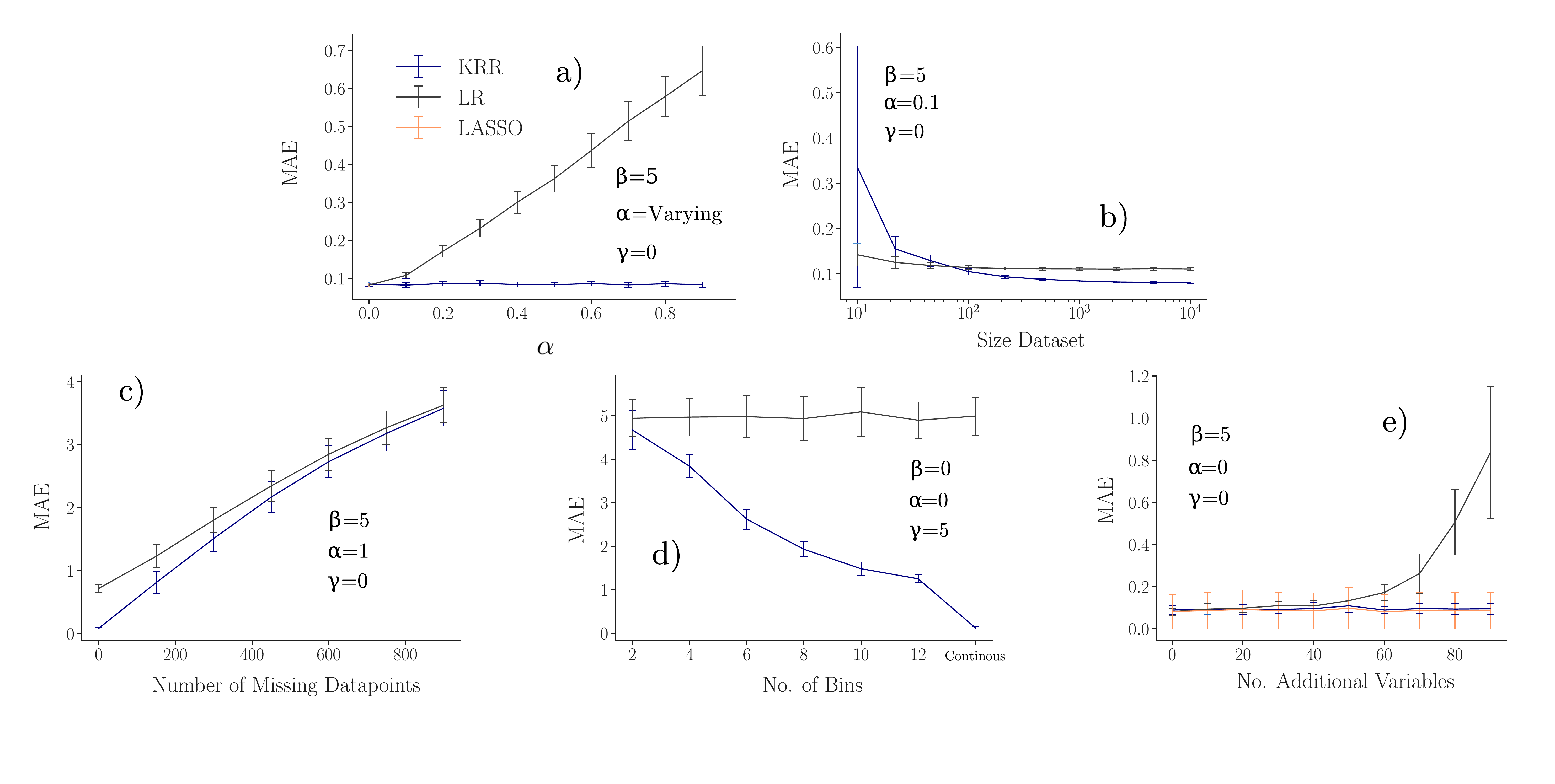}
    \caption{ The change in performance of a linear (LR) and kernel ridge regression (KRR) model with properties of the data. The simulated data has the form: $ \small f(\mathbf{x}) = \beta_0 x_0 + \beta_1 x_1  + \alpha x_0 x_1 + \gamma x_0^2  + \epsilon $ where $\epsilon \sim N(0.5,0.1)$ and the coeffients $\alpha$, $\beta$ and $\gamma$ vary throughout the examples. The difference in error: a) with the $\alpha$ used in the simulation b) the size of the dataset, c) the number of missing points for the continuous variable, d) the number of bins used for the continuous variable and e) the number of variables present for the simulated dataset. For part e), the additional variables added do not contribute to $y$ but are correlated with $x_0$ and $x_1$. This results in a decrease in the performance of the unregularized linear model. LASSO is a linear regression technique with an additional regularization term in the loss function to prevent overfitting. MAE is the mean absolute error, a measure of the accuracy of the model.}
    \label{fig:Sim_model}
\end{figure*}

To demonstrate the different factors that affect model performance described in this section, we have used simulated datasets to compare the difference in performance between an interpretable linear model (with the simplistic form $f(\mathbf{x}) = \beta_0 x_0 + \beta_1 x_1$) and kernel ridge regression (KRR). This is shown in Fig.~\ref{fig:Sim_model}. Kernel ridge regression uses a non-linear kernel alongside regularized linear regression. This allows functional forms to be recreated that have not been explicitly included in the functional form.  As characteristics, such as the size of the dataset, are varied a change in the relative performance of the linear model and kernel model can be seen. We will refer to this figure throughout our discussion.

\subsection{The Problem is Linear and Additive}
One reason that non-linear methods will not improve over a linear model is because the underlying problem is linear and additive. Before embarking on a project it is useful to ask: is there a reason to suspect non-linearity is present in the problem? In other words, how will a flexible functional form help the predictive performance of this problem? If the relationships between the independent and dependent variables are known to have linear relationships, and interactions between variables are not expected, then it is unlikely that improvements will occur. However, it is worth noting that a properly regularized, well-trained, non-linear model should not perform any worse than a linear model for this case.

Even if the problem itself is non-linear, creating a linear model can serve as a baseline for comparison. This can shown the improvements that occur in comparison to a linear functional form.

\subsection{Transforms and Interactions Can Be Explicitly Added to a Linear Model}
For certain problems, the relationship between the independent and dependent variables is approximately known or can be inferred with an ``educated guess'' -- often based on prior domain knowledge -- and can be modelled by a simple functional form. In these cases, using a linear model with transformed variables can result in an accurate predictive solution and the flexible functional form of ML models is not necessary. Similarly, a class of interactions may be hypothesized and added to the model.  

\begin{figure}[ht!]
    \centering
    \includegraphics[width=3.50in]{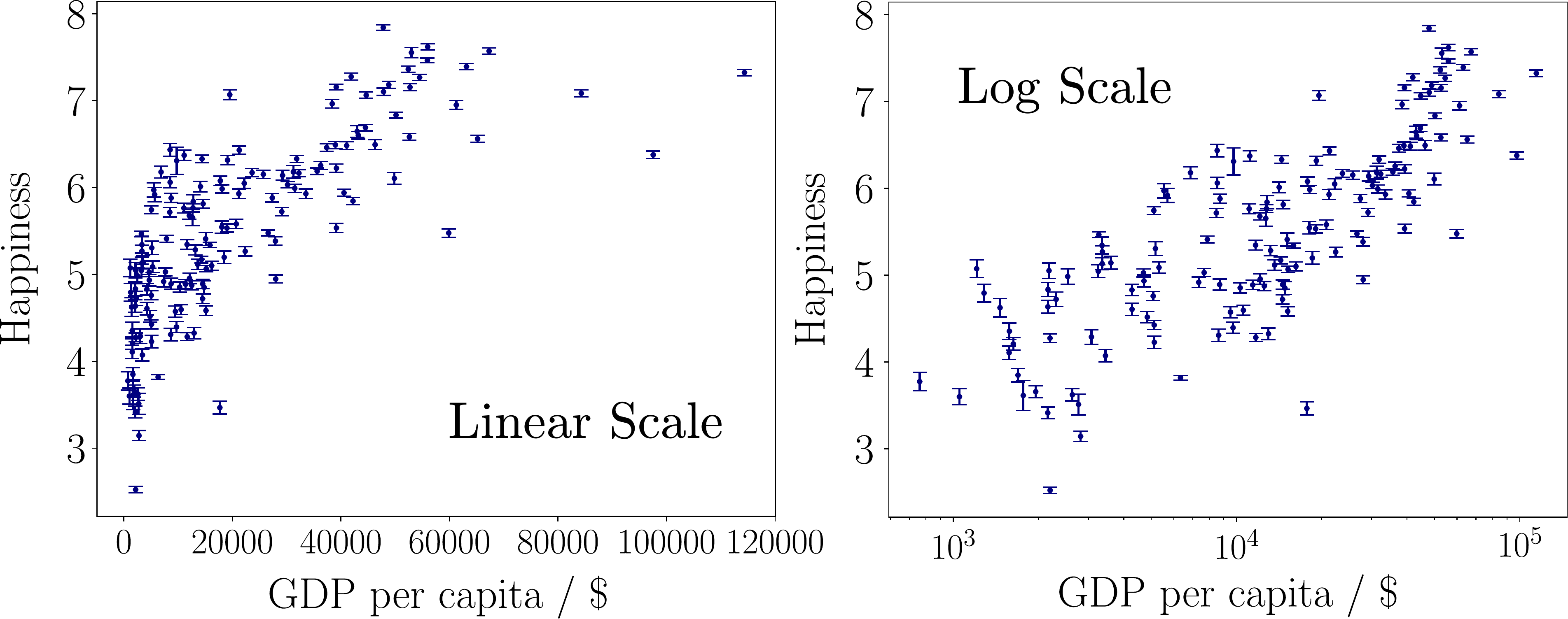}
    \caption{The relationship between reported happiness and gross domestic product (GDP) per capita of a country. The data is taken from Ref.~\citenum{Helliwell2021}. On a log scale a clearer linear relationship between the GDP per capita and happiness can be seen than on a linear scale. The transformed variable could then be used directly in a linear model.   }
    \label{fig:log_income}
\end{figure}

For example, in predictive models of life satisfaction the logarithm of income is often taken. The reasons for this can be seen on a countrywide scale in Fig.~\ref{fig:log_income}. Interactions can also be hypothesised, for example in Ref.~\citenum{Strand2014} interactions between ethnicity and socio-economic status were added to linear models for educational achievement.

Alternatively, there are sets of problems which can benefit from using automated techniques, such as symbolic regression, to search for linear combinations of non-linear basis functions~\cite{Wang2019_sr, Udrescu2020, Ouyang2018}. Symbolic regression automatically searches for an analytical formula to describe a relationship and this can provide a highly interpretable solution~\cite{Udrescu2020}. This is not suitable for all problems due to computational scaling restrictions but is an excellent route for some applications. 

\subsection{Binary and Categorical Variables}
\begin{figure}[ht!]
    \centering
    \includegraphics[width=2.5in]{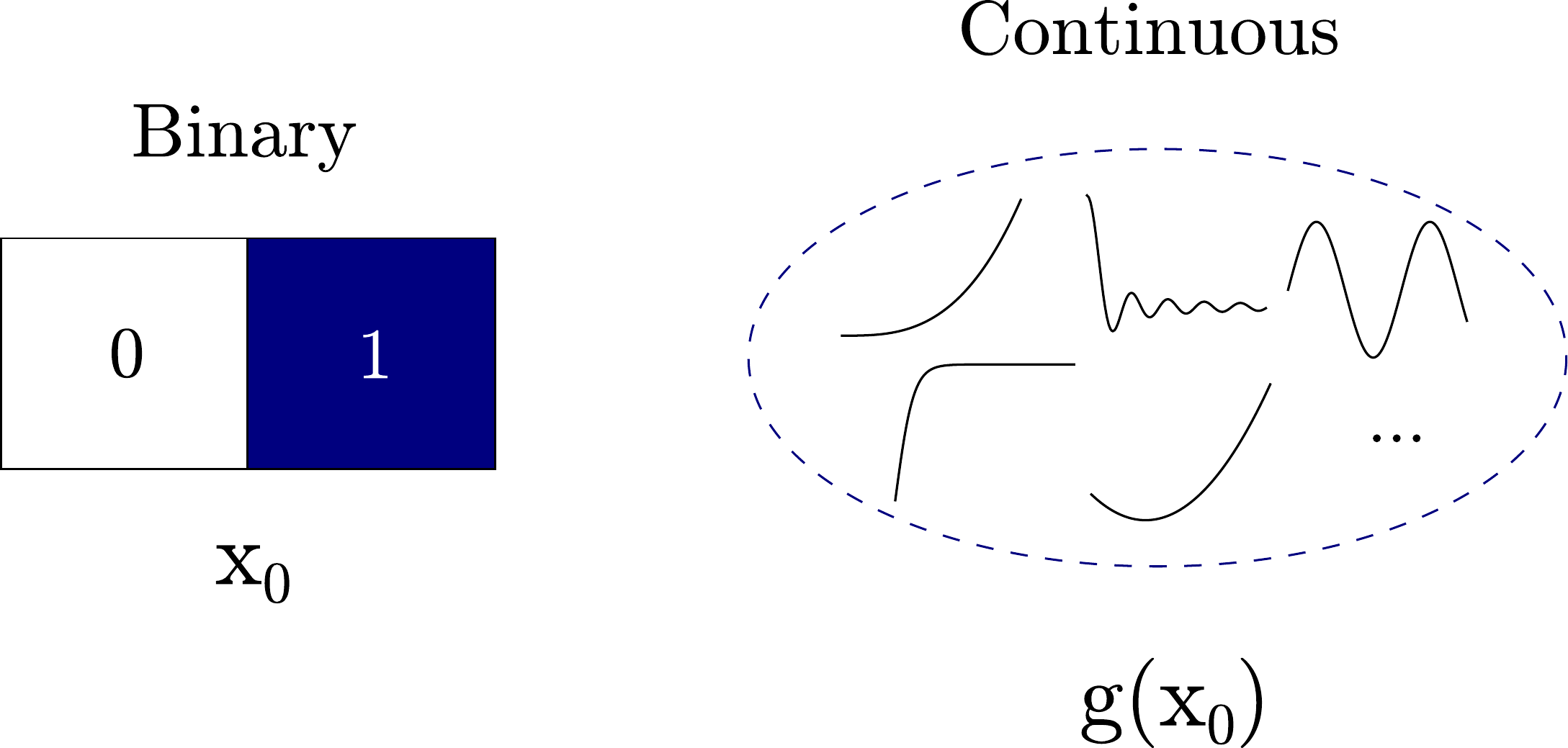}
    \caption{The difference in the space of possible solutions for a binary variable and a continuous variable. For a binary variable, the space of all possible transforms does not need to be considered. }
    \label{fig:Bin_con}
\end{figure}

Binary, or categorical/ordinal variables that can be one hot encoded or used as dummy variables, cannot exhibit non-linearity except by interaction terms between variables (see Fig.~\ref{fig:Bin_con}). Therefore, if a large number of the variables present only take binary values the possible routes for improvements upon linear models are limited. Additionally, it is straightforward to manually improve linear models if only binary variables are present as the set of possible solutions is finite. For example, if $x_0$ and $x_1$ are binary then $y=ax_0 + bx_1 + cx_0x_1$ would describe all possible states.

\subsection{Non-linearity is Present But Only Slightly}
It is possible that non-linear relationships do exist but the variables concerned have a small contribution to the outcome. In Fig.~\ref{fig:Sim_model} a), the performance for a linear and kernel model is compared as the non-linear contribution increases. As the strength of the interaction grows, the performance of the linear model decreases. This is because the linear model does not include an interaction and has a limited functional form. It may also be that the non-linearity present is masked by noise. For $y=f(x) + \epsilon$ if $\epsilon$ is very large the model may be unable to discern the underlying $f(x)$.

Alternatively, non-linearity may be present only at the extremes of the distribution (where few points exist). Slight non-linearity which does not influence predictive performance is particularly likely if there are many variables contributing to the outcome.

As an example, let us imagine that we are building a predictive model for a child's test score at age 12, and this test score has a highly linear relationship with the measure IQ of the child except for the top and bottom 2.5$\%$ of the score. In this region, it departs from linearity. Using a non-linear model may help to account for this deviation, however if 95$\%$ of the data can be described by a linear model then the improvements in predictive accuracy with a non-linear model may be negligible, even though slight non-linearity does exist.

\subsection{Error in the Independent Variables}
\label{sec:error_inv}
Improvements may not occur due to problems with measurement (i.e. binned measures of continuous variables) or processing (i.e. replacement of the missing values by the mean values) of the independent variable. This can lead to non-linearity present in the underlying relationship being lost. The difference in performance when a continuous variable is placed in discrete bins is shown in Fig.~\ref{fig:Sim_model} d). When no bins are used the KRR model is more accurate than the linear model, but as successively more bins are added the difference in performance is reduced. Alternatively, if there are missing values present in the dataset this can also reduce the improvements seen with non-linear methods. In  Fig.~\ref{fig:Sim_model} c), the difference in the linear and KRR model shrinks as the number of missing data-points grows. The presence of error in the independent variable has been described as a distractor variable in Ref.~\citenum{Wilming2022}.

For example, consider that we are collecting data by asking someone to state their weight within 5 different ranges ($<$30kg, 30-50kg, 50-70kg, 70-90kg, $>$90kg). Information is lost as the values have been binned. There may be other problems with the data, for instance the person may be entirely wrong about the weight they enter. Additionally, some people will miss the question entirely, leading to the need to impute their values in preprocessing steps. All of these factors can reduce the improvements seen with non-linear methods. 

\subsection{The Dataset is Too Small}
Even if the underlying relationship between the independent variables and a predicted variable is complex, if there is only a small dataset then the non-linear model will not be  able to recreate this. The differences between a linear and a non-linear model may only become apparent at a larger dataset size. This is demonstrated in Fig.~\ref{fig:Sim_model} b), where the superior performance of KRR is only seen when the dataset is sufficiently large. Complex non-linearity in a model is not useful if the dataset is not large enough to exploit it.

For interactions between discrete, categorical variables the limitations with the dataset size can be more clearly seen. Consider a problem which contains 4 independent discrete, categorical variables that can take $k$ possible values ($x_0 = $1, 2, 3 ... $k$) with each value having equal probability of occurring. Approximately 1/$k$ of the dataset will have a value of $x_0=1$. Therefore, if $k=10$ only 10\% of the data will contain $x_0=1$. For pairwise interactions (ie. $x_0x_1$), the percentage of the dataset containing a specific interaction will drop rapidly to only 1\% of the dataset. This trend would continue and only 0.01\% of the dataset would contain a specific interaction between all four variables. Therefore, to accurately describe high order interactions the dataset would have contain tens of thousands of data points. 

\subsection{When Interpretability is More Important Than Accuracy}
Statistical models are important not only for predictions but also for causal explanations~\cite{Shmueli2010, Prosperi2020,Pearl2010}. In many disciplines, being able to derive causal explanations from a model is more important than predictive accuracy~\cite{Shmueli2010}. Causal explanations from ML models is an active area of research and this is not a trivial step ~\cite{Lecca2022,Prosperi2020,Chattopadhyay2019}. Therefore, if causal explanation is required, it may be preferable to create statistical models using traditional techniques. However, drawing causal conclusions from statistical models does come with its own complications - in part due to their restrictive functional forms~\cite{Hainmueller2019, Brambor2006}. Therefore, with the growth of causal techniques for non-linear methods these techniques may yet become the preferred choice~\cite{Runge2019}. 

Interpretability can have other advantages. For material science applications, the construction of interpretable models by automated searches has clear advantages. In Ref.~\citenum{Bartel2019}, automated searching was used to find an expression  describing if an ABX3 materials was perovskite or nonperovskite. The expression found is shown below: 
\begin{equation}
    \tau = \frac{r_X}{r_B} - n_A(n_A - \frac{r_A/r_B}{ln(r_A/r_b)})
\end{equation}
This equation has a 92\% accuracy for dataset of 576 experimental structures. It is likely that accurate predictive models could have been found with ``black box'' methods, however the well defined formula allows general behaviours to be seen and is a transparent solution~\cite{Bartel2019}.

\section{When Non-linear Models are Required}
On the other hand, there are many reasons why improvements may occur with a non-linear ML model and they are an essential tool for many complex problems.

\subsection{Intricate Non-linearity is Present}
The primary reason for improvements occurring with non-linear models is due to non-linearity being present in the underlying relationship that has not been accounted for in a linear model. Such intricate non linearity might be, for example, an interaction between three or more independent variables. A non-linear ML model is able to recreate complex relationships due to its flexible functional form and this can lead to dramatic improvements in performance. For certain applications in the sciences, knowledge can be incorporated into a model to improve performance through manually engineered features~\cite{Muller2003}. Unfortunately, manually engineered features cannot be found for all problems and therefore flexible functional forms must be employed.

\subsection{Regularization}
 Overfitting happens when a complex model too closely recreates the training data, and in doing so does not generalize to data outside of the training set. The accuracy of a ML model is evaluated by splitting a dataset into a training and testing set. The training set is used to fit the model, and the testing set is used to evaluate the performance of a model. Training and testing sets are necessary for ML models to ensure overfitting is not occurring. Overfitting can occur for linear models, but only when the number of variables is high compared to the size of the dataset and the data is correlated, and this will be seen from the standard errors of the coefficients. Regularization prevents overfitting occurring and is another reason that non-linear ML models may lead to improved accuracy over an unregularized linear, additive model.

For linear models, adding a penalty term to the loss function, ie. LASSO, ridge or elastic net regression is a form a regularization~\cite{Tibshirani1996, Hoerl2000, Zou2005}. For non-linear models, regularization can also take the form of a penalty term in a loss function, or can be provided for by alternative methods, such as dropout approaches for NNs~\cite{Srivastava2014}. 

If a problem has a large number of variables and a comparably small dataset, non-linear models may show improvements over an unregularized linear model even when the underlying relationship contains no non-linearity. This is because regularization is preventing overfitting from occurring. This is demonstrated in Fig.~\ref{fig:Sim_model} e) where the performance of the unregularized linear model decrease as more variables are added to the dataset. In contrast, the regularized KRR model and LASSO model do not decrease in performance.

\subsection{Flexibility in the Solution}
An adaptable functional form can have additional advantages. A model may be required that can easily be used for a large range of different datasets or can adapt to changes as more data is added. If the main priority is producing accurate predictive models quickly, searching for intrinsically interpretable models may not be a constructive use of time and post-hoc interpretability techniques could instead be used to analyse the models produced.

\section{Interpretability Techniques for Machine Learning Models}
To visualize the non-linearity present in real datasets, we use two interpretability techniques: partial dependence (PD) and individual conditional expectation (ICE). These particular post-hoc interpretability are mentioned as they are used in the following sections, for a more complete review of post-hoc interpretability methods, see Ref.~\citenum{Samek2021} or Ref.~\citenum{Du2019}.

The relationships present in a  model can be visualized with partial dependence, which is defined as:
\begin{equation}
PD(x_s) = \frac{1}{n} \sum^n_{i=1} [\hat{f}({x_s, x^{(i)}_c})] 
\label{equ:partial_dep}
\end{equation}
where $x_s$ is the variable of interest, $x_c$ is all other variables excluding $x_s$ and $\hat{f}$ is the ML model. The PD is a measure of the mean value of the ML model at $x_s$ across a dataset. 

From equation \ref{equ:partial_dep}, we can see that partial dependence averages over multiple contributions of $\hat{f}({x_s, x^{(i)}_c})$. This averaging step can hide important information such as the presence of interactions between variables. Individual Conditional Expectation (ICE) plots show all points $\hat{f}({x_s,x^{(i)}_c})$ for a given $x_s$ and this avoids the issues associated with averaging over the points. 

In a predictive model, interactions between variables may also exist. If a function, $F(x)$, where $x = (x_1, x_2, ..., x_n)$, cannot be expressed as two functions:

\begin{multline}
F({x}) = f_{\setminus j}(x_1, ..., x_{j-1}, x_{j+1}, ..., x_n) \\   + f_{\setminus i}(x_1, ..., x_{i-1}, x_{i+1}, ..., x_n)
\end{multline}

where $f_{\setminus j}$ does not depend on $x_i$ and $f_{\setminus j}$ does not depend on $x_i$ then an interaction is said to exist between variables $x_i$ and $x_j$~\cite{Sorokina2008}. Many different strategies have been developed for identifying the interactions present in a model~\cite{Apley2020, Sorokina2008, Hooker2007, Strumbelj2013, Ribeiro2016, Lundberg2017, Friedman2008}. This includes using partial dependence in the form of H statistics. The interaction between two variables $j$ and $k$ is defined by:
\begin{equation}
\small
H^2_{jk} = \frac{ \sum^n_{i=1} [PD_{jk} (x^{(i)}_j, x^{(i)}_k) - PD_j(x^{(i)}_j) - PD_k(x^{(i)}_k)] }{ \sum^n_{i=1} PD^2_{jk} (x^{(i)}_j, x^{(i)}_k)}
\end{equation}
with $H_{jk}=\sqrt{H^2_{jk}}$. For meaningful interpretation, the H statistic for $\hat{f}({x})$ has to be compared to the distribution of H statistics if no interactions are present in the underlying model. The creation of a null distribution is discussed in further detail in Ref. ~\citenum{Friedman2008}. The code used to calculate and produce the figures in the following section is provided (https://github.com/aa840/icepd).

\section{Multidisciplinary Applications}
We have outlined several reasons why non-linear methods will and will not improve predictive performance, and demonstrated this with simulated datasets. We will now look at examples and, in doing so, explore why improvements in performance do or do not occur. Furthermore, we will also introduce perspectives from multiple disciplines. Drawing from multiple perspectives, we will demonstrate the benefits that still remain even when improvements in predictive performance are not seen and highlight the advantages of large, complex linear models that contain thousands of basis functions. The examples and insights obtained are summarized in Fig.~\ref{fig:Summary_diag}.

\subsection{Social Sciences - Predicting Height}

We begin by looking at a predictive model for height in adulthood (at age 34) using variables describing physical and socio-economic conditions in childhood (at age 10). In doing so, we will show how non-linearity can be clearly present in a NN model but if there is not a high density of data in the region with non-linearity then improvements in predictive performance are unlikely to be seen.  
The dataset used in this example is from the British Cohort Study (BCS), a longitudinal cohort study of around 17,000 children born in 1970 in England, Scotland and Wales~\cite{Sullivan2022}. A NN model was built that predicted the body mass index (BMI) of the participant at age 34 using information about the parents and child when the child was age 10.

\begin{figure*}[htb]
    \centering
    \includegraphics[width=6.75in]{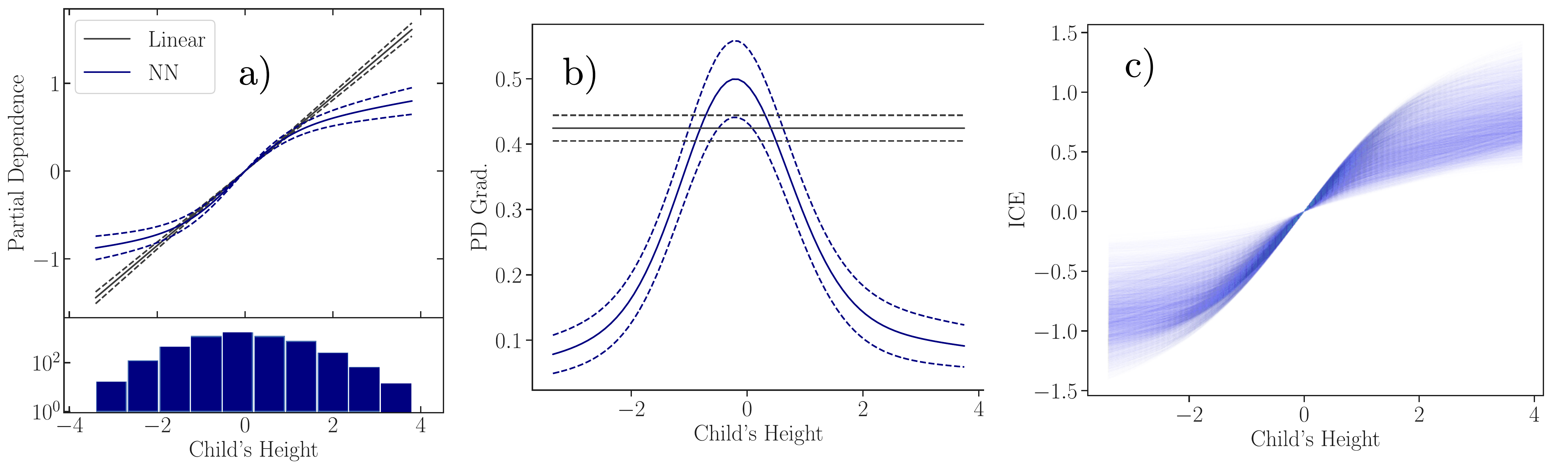}
    \caption{For the NN model for predicting adult height, the a) PD plots, b) gradient of PD and c) ICE plot for child's height are shown. Clear non-linearity can be seen at the extremes of height.  }
    \label{fig:Height_PD}
\end{figure*}

The following variables were used: Gender, Child's Height/Weight, Father's Height/Weight, Mother's Height/Weight, Own House (do the parent's own their house, a measure of wealth), Parent's Income per Week. The total number of number of data points used is 6283 and all non-binary variables were standardized to $\mu = 0, \sigma^2 = 1$. The number of variables is restricted for simplicity.

The linear and NN model have comparable accuracy, with R$^2$ score of 0.77 and 0.78 respectively. The partial dependent plots for both model are generally similar, as seen in Fig. S2. However, for the  child's height at age 10 a deviation from the linear solution is seen at the extremes of weight with the gradient decreasing past one standard deviation from the mean value - see Fig.~\ref{fig:Height_PD}. The distribution of variable is shown underneath the figure and highlights the lack of data in the regions where the NN deviates most strongly from the linear solution. The gradient of this partial dependence plot is shown in Fig.~\ref{fig:Height_PD}b) to highlight departures from the linear solution and the ICE plot shows the variation in the relations across data-points. 

The interactions present are considered in  Fig. S1. The H-statistics for the NN-model shows only one interaction, between the Mother's weight and the gender of the child and the ICE plots for the mother's weight are shown in Fig. S3. Therefore, there are not a large number of interactions present in the model.  

By examining the NN model for BMI prediction using interpretability techniques, we can now draw conclusion about why the linear and non-linear models have similar performance. Non-linearity and interactions do exist in the NN solution but only at extremes of the distribution and the strength of the interaction between variables is weak. Further reasons exist for the comparable performance due to the nature of the dataset. Firstly, binning is used for the family income. Income is originally a continuous variable but it is then placed into discrete bins and this can cause a deterioration in predictive performance as discussed in Section \ref{sec:error_inv}. Additionally, missing values in the dataset are replaced by mean values and this will also prevent higher accuracy levels. Despite the lack of improvement, the use of NN models may still be beneficial as discussed in the following section.

\subsection{Political Sciences - Identifying Non-linearity in Interaction Models}

Interactions are often added to linear models to test conditional hypothesis, for example in Ref. \citenum{Strand2014} interactions were used to study how factors related to class, gender and ethnicity combine in a non-additive way to impact educational achievement. However, if non-linearity is not properly described in a model this can lead to spurious interactions being recognised~\cite{Hainmueller2019}. ML combined with interpretability techniques provide a unique method to describe non-linearity in a model and to see if interactions remain when this non-linearity is properly described. 

In Ref.~\citenum{Hainmueller2019}, the various problems associated with adding interactions to linear models to draw causal conclusions were discussed and numerous datasets revisited. The datasets involved many topics including  sustainable energy transition, the role of central bank independence and the electoral benefit of failed no-confidence motions~\cite{Aklin2013, Bodea2015,Williams2011}. This analysis included using kernel models, which have a flexible functional form, to investigate whether an interaction between $x_0$ and $x_1$ was linear with respect to both variables ($x_0x_1$) or if non-linearity existed in one of the terms ($g(x_0)x_1$ where $g(x_0)$ is a non-linear function of $x_0$). Incorrectly assuming linearity for the terms in the interaction can result in statistically significant interactions being found that are no longer statistically significant when the non-linearity has been corrected for. This is problematic when modelling is used for causal explanations. This demonstrates how non-linear models can be used for purposes other than just improving predictive performance. 

\subsection{Biology - Small Interfering RNA Example}
The discovery of small interfering RNA (siRNA) and its role in gene regulations has been a key development in  molecular biology~\cite{Fire1998, Dana2017}. By interacting with messenger RNA (mRNA), siRNA can prevent translation occurring, interrupting the production of amino acids and regulating gene expression~\cite{Dana2017}. Large scale siRNA screens are regularly used to better understand biological process and, due to siRNA's ability to induce gene silencing, it is also seen as a potential therapeutic tool with multiple siRNA therapeutics recently approved~\cite{Dana2017, Hu2020, Hadjslimane2007}. 

The effectiveness of the gene silencing by siRNA can vary considerably~\cite{Khvorova2003, Walters2002, Amarzguioui2004}. Therefore, multiple attempts have been made to build predictive models of the inhibitory activity of siRNA~\cite{Reynolds2004, UiTei2004, Amarzguioui2004, Shabalina2006, Peek2007, Sciabola2013,  Sciabola2021, Huesken2005, McQuisten2009, Ichihara2007, He2017}. siRNA is made up of four bases, adenine (A), cytosine (C), guanine (G) and uracil (U) joined in a sequence, with a typical length of around 21 base pairs. This sequence information can be used as features for a predictive model~\cite{Huesken2005, McQuisten2009, Ichihara2007}. Alternatively, the frequency of bases and neighbouring bases (ie. the number of A, AC, etc.) can be used~\cite{He2017}. Secondary structure, thermodynamic information or mRNA related features can also be incorporated into a model and this has been shown to result in improvements in performance~\cite{He2017, Sciabola2013, Khvorova2003, Shabalina2006}. We will focus on positional sequence features of siRNA, as these features have been shown to offer competitive performance and are regularly included in predictive models of inhibitory activity of siRNA~\cite{McQuisten2009}. 

Before the development of regression models for predicting inhibitory activity, rules based methods were proposed~\cite{UiTei2004, Reynolds2004,Amarzguioui2004}. These methods described simple criteria, such as lower G/C content or a specific base present at a given position, that were associated with higher efficacy~\cite{Reynolds2004}. Subsequently, linear models, artificial NNs and support vector machines have all been proposed as predictive models for inhibitory activity. One of the first ML models for predicting siRNA inhibitory activity was BIOPREDsi~\cite{Huesken2005}. However, it was later seen that a linear model, $i-Score$, showed comparative accuracy to BIOPREDsi and also just relied on positional sequence information~\cite{Ichihara2007}.  A detailed comparison of each model's relative performance with varying feature sets is given in Ref.~\citenum{McQuisten2009}.

In this work, we fit a NN model that uses sequence information to predict the inhibitory activity. The predictions from this NN model are highly correlated with the BIOPREDsi model, $\mathrm{R}^2$ of 0.97. By employing ML interpretability methods, the NN model is shown to effectively recreate the linear solution, with near identical PD values. We will discuss the reasons why this is the case. The dataset used is from Ref.~\citenum{Huesken2005} and contains 2431 siRNA sequences  provided along with experimentally determined inhibitory activity. Details of the dataset and model used is given in Section S1.2.

The linear model and NN model for the positional sequence have $\mathrm{R}^2$  of 0.43/0.42 and 0.45/0.53 for the training and testing set respectively. This is similar performance to that seen in Ref. ~\citenum{Ichihara2007}. We can now further investigate the NN solution using interpretability methods. 

\begin{figure}[ht]
    \centering
    \includegraphics[width=3.00in]{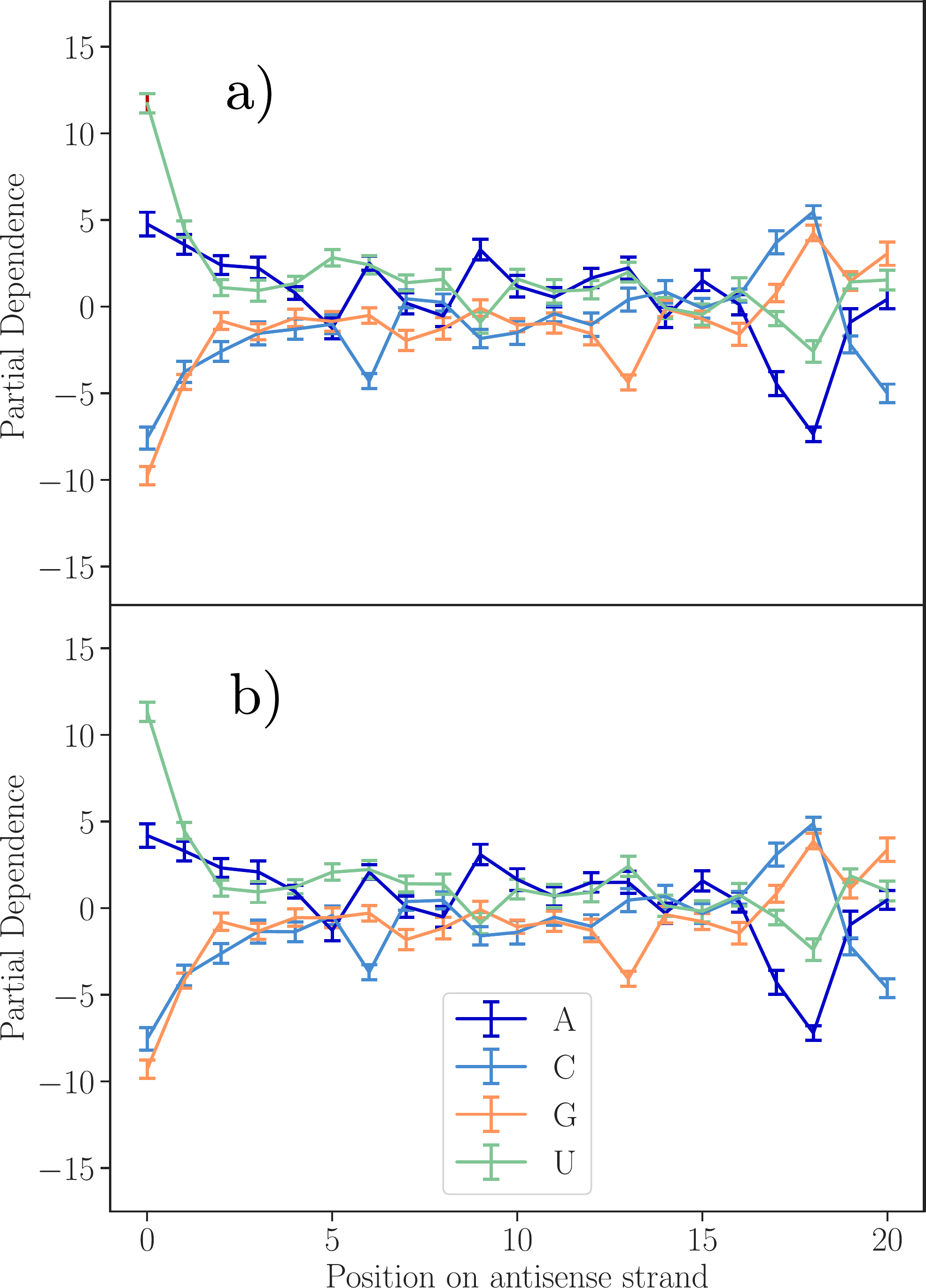}
    \caption{The a) partial dependence of a linear model and b) the partial dependence of a NN model fit to the siRNA sequence data. The mean for the partial dependence plot is set to be equivalent to the linear plot. The error bars show the standard deviation. }
    \label{fig:PD_linear_NN}
\end{figure}

Figure ~\ref{fig:PD_linear_NN} shows the partial dependence of the linear model alongside the partial dependence of the NN model. Two considerations should be noted, firstly as the variables in this example are binary, non-linearity cannot occur in model except through interactions. Secondly, the coefficients and partial dependence of a linear, additive model differ just by a constant. The partial dependence for the NN and linear model are almost identical with the difference in the partial dependence between the two models averaging around 0.2. Therefore, there is a negligible difference in the PD between the two models. 

\begin{figure}[ht!]
    \centering
    \includegraphics[width=3.10in]{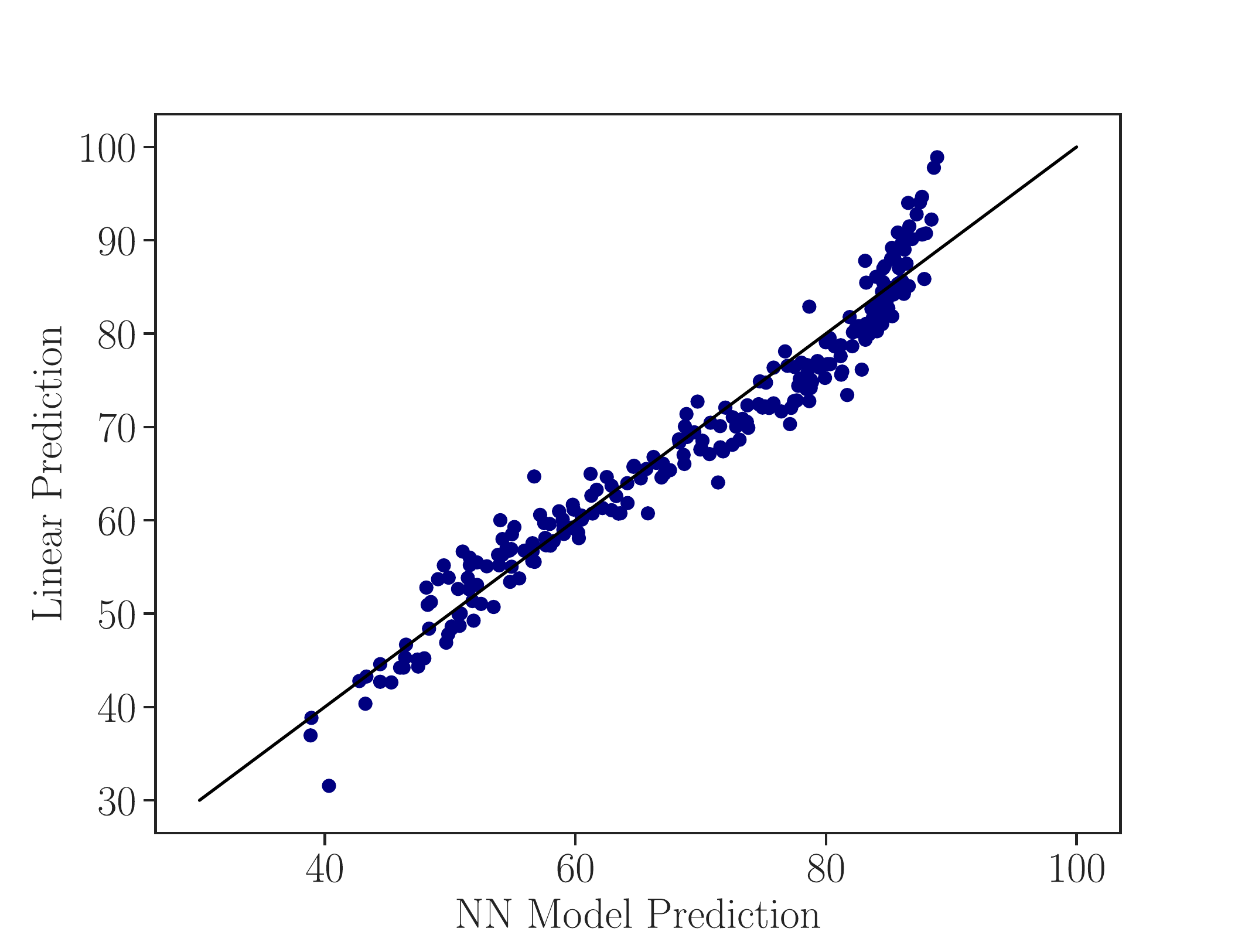}
    \caption{The predicted results for the siRNA inhibitory activity for the test set for the linear and NN model. The black line is at $y=x$.}
    \label{fig:pred_NN_lin_test}
\end{figure}

The similarity between the two models can also be assessed by comparing the predictions for the test set with the linear and NN model, Fig.~\ref{fig:pred_NN_lin_test}. From this graph, it can be seen that the two predictions are highly correlated. Interactions in a model can also be used to compare the model and can be assessed through a variety of methods~\cite{Strumbelj2011, Friedman2001, Goldstein2015, Hooker2007, Sorokina2008}.  Whilst interactions do exist in the NN model, see Fig. S4 where H-statistics are used to assess interaction strength, these interaction are not strong enough to cause large changes in the NN model's prediction.

For this example, we have seen comparable performance between a linear and non-linear model. This is partly because the problem contains binary variables. As previously discussed, binary variables can only have interactions between terms and interactions can easily be added to a linear model manually if there was a significant difference in performance. Another reason for the lack of improvement is the size of the dataset - there are 2431 datapoints and 84 variables. There are over three thousand possible pairwise interactions that could exist, and the sampling of these interactions with the limited size dataset is poor. Therefore, even if there were strong interactions present in the underlying problem, the sampling is insufficient to capture a significant amount of them.

\subsection{Material Sciences - Prediction of Formation Energy in Crystalline Materials}
In this discussion so far, we have focused on examples where linear models provide approximately the same accuracy as non-linear models. For this example, we turn to a case where complex non-linear solutions are required and the manual addition of interactions and transforms is not possible. ML models can provide fast and accurate predictions of material properties, and we will look at the prediction of formation energy for crystalline materials. This is a problem that requires non-linear solutions and we will explore the non-linearity present with interpretability methods.

The predicted quantity in this example is formation energy which describes the stability of a material. The features used are the stoichometric formula, which describe the percentage of each element in a material. For example, $Al_2O_3$ would have features Al=0.4 and O=0.6. This is a very simplistic feature as it does not contain information about distances between atoms or the structure of the material. It is because these features are so simple that complex non-linearity is required for the regression model. 

A dataset of crystalline materials was taken from Ref.~\citenum{Faber2015}. A NN model with six hidden layers was created for predicting the formation energy of a material with stoichometry used as the input variable. The stoichometry defines the percentage of each element present in a material. This methodology is similar to the used in Elemnet, from Ref.~\citenum{Jha2018} although the NN architecture is not identical. Stoichometry has also been used with a NN framework with the Roost package~\cite{Goodall2020}. 

\begin{figure}[htb]
    \centering
    \includegraphics[width=3.20in]{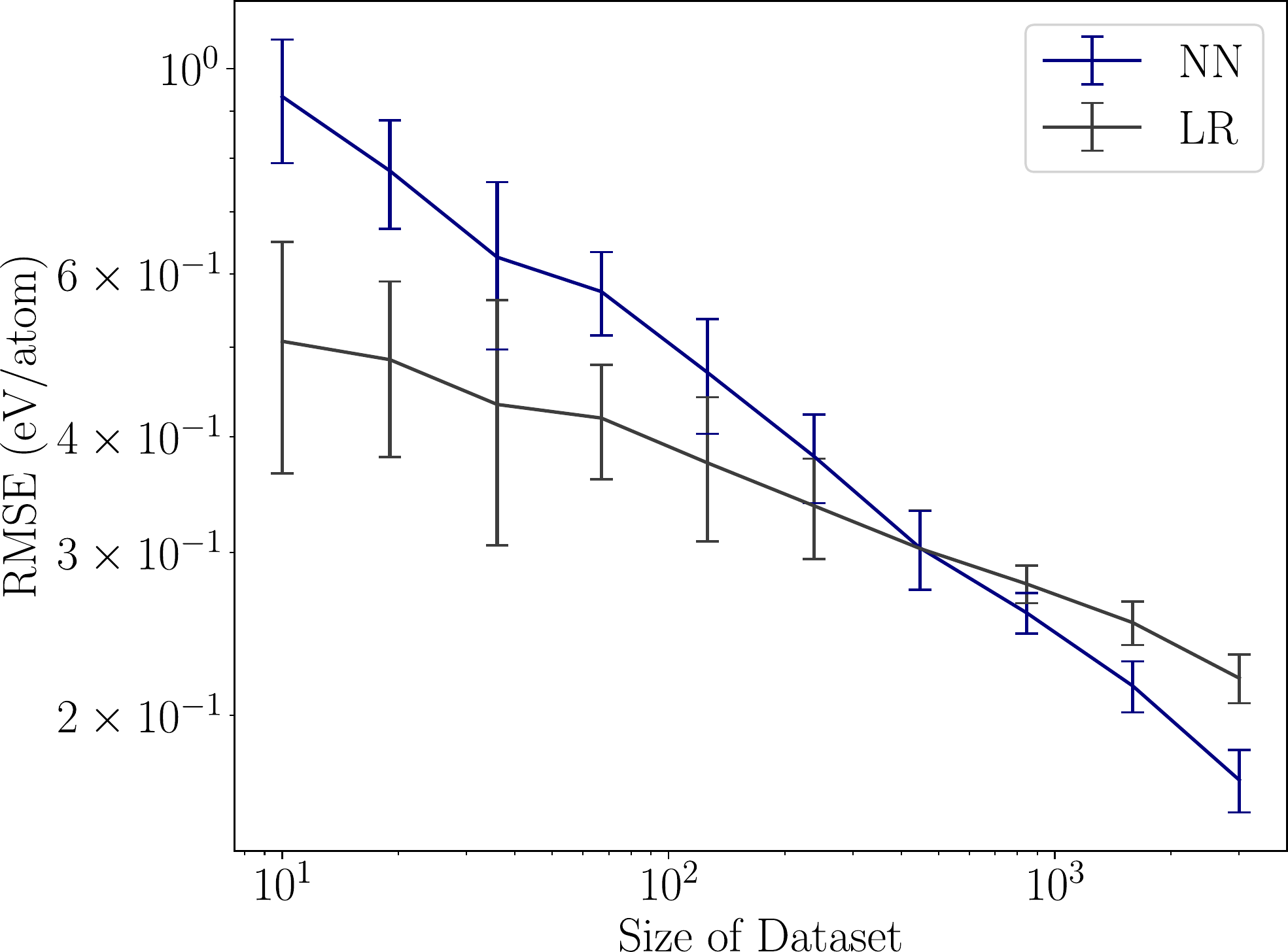}
    \caption{The change in error with the size of the dataset used for the NN and linear model. As the dataset size increases, the NN begins to outperform the linear model. The error bars show the standard deviation for different train/test splits. }
    \label{fig:LC_Fe}
\end{figure}

The learning curve of the NN model and a linear model is shown in Fig. \ref{fig:LC_Fe}. The stoichometric information is a very limited set of information to provide and so complex non-linearity must be present for the accurate prediction of formation energies. The linear model cannot compete as this simplistic model does not capture the chemical behaviour. 

\begin{figure}[ht]
    \centering
    \includegraphics[width=3.20in]{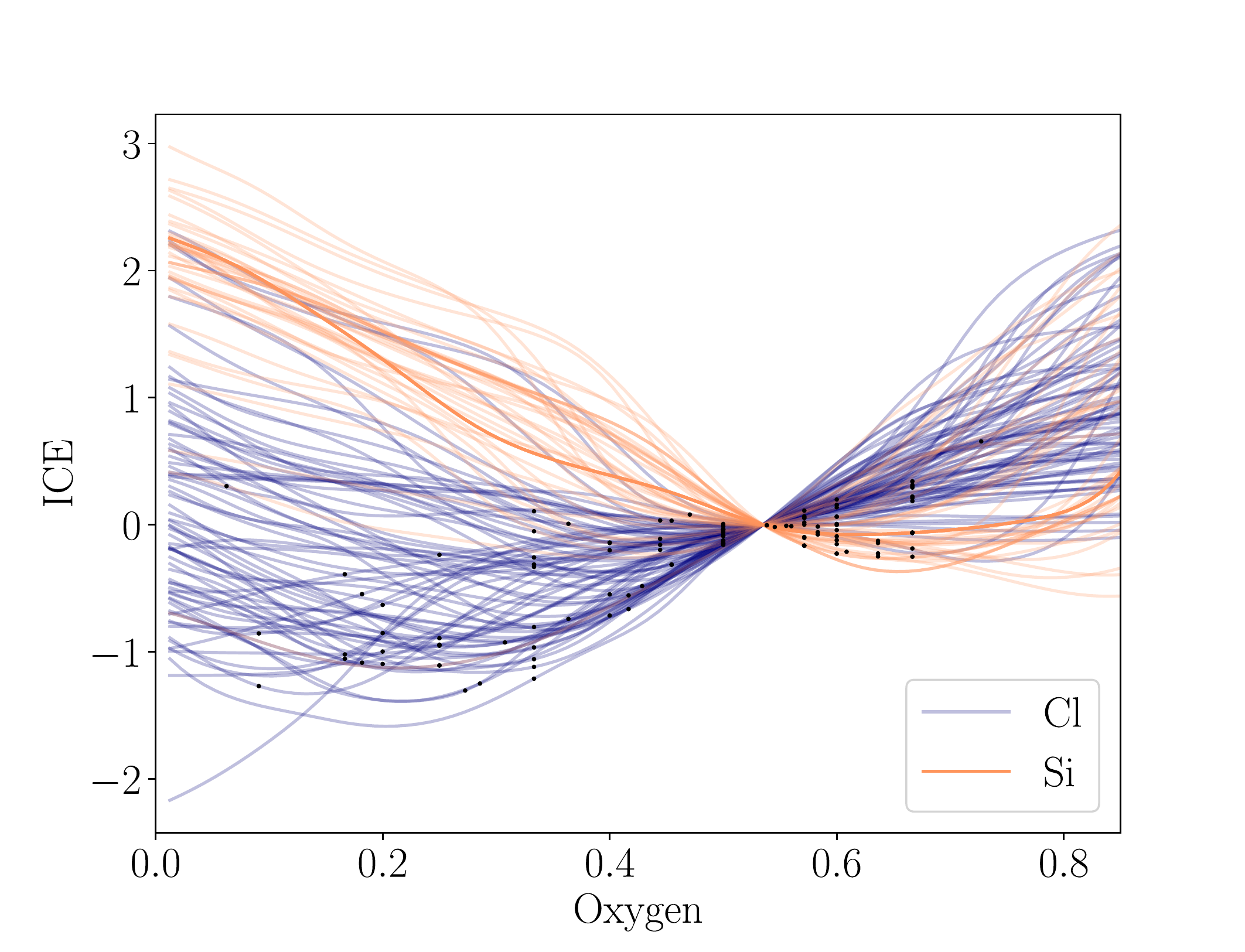}
    \caption{The ICE plots for oxygen for materials containing silicon and chlorine. The data-points present in the training set are shown by black markers. Clear non-linearity in the solution can be seen. }
    \label{fig:ICE_FE}
\end{figure}

The non-linearity present in the NN solution can be explored with ICE plots. These are shown in Fig.~\ref{fig:ICE_FE} with the change in the prediction with oxygen content given for chlorine and silicon containing materials. The varying gradient of the ICE lines indicates that the relationship between the percentage of oxygen and the formation energy predicted is dependent on other variables present in the material. In other words, interactions are present in the model.  Furthermore, the different relationships for chlorine and silicon indicate that linear-additive relationships would not suffice to describe the effect that oxygen concentration has on formation energy, Fig.~\ref{fig:ICE_FE}.

For this example we have explored the reasons why NNs improve performance over simplistic linear model for formation energy prediction. In doing so, we have provided an example of why non-linear methods do improve predictive performance and shown the clear non-linearity present in the model. 

\subsection{Chemistry - Linear Interatomic Potentials}
\label{sec:chem}
Before we conclude, it is important to highlight that complex non-linearity can be represented by a linear model. If we have a very large number of possible non-linear basis functions (with varying transforms and interactions present) this is capable of modelling highly complex non-linear problems. This has been seen in applications to chemistry and materials science. 

Interatomic potentials are used in computational chemistry and material science to predict the forces and energies present in a system~\cite{Shapeev2016, bartok2010gaussian, Keith2021, Deringer2019-ss, sGDML}. Given a set of training data containing the coordinates, energies and (optionally) forces for a molecular or material system a predictive model can be built. Whilst simplistic classical models have been used for these problems, greater complexity is required to fully model the interactions present in an atomistic system~\cite{Jorgensen1988,Weiner1984}. ML models have been used as interatomic potentials for over ten years but have traditionally consisted of kernel or NN models~\cite{Smith2017, Schutt2017, bartok2010gaussian, Behler:2011it, ChemRev-MLFF}. In recent years it has been shown that complex linear combinations of non-linear basis functions can also provide models that are both fast and accurate ie. (MTP, ACE, SNAP, ChIMES etc.)~\cite{Drautz2019, Shapeev2016, chimes1, Van_der_Oord2019-td, Allen2021, Kovacs2021}. 

Models that use linear or polynomial regression with a very complex set of basis functions and high order interaction terms are an alternative approach for describing complex non-linearity in a system. For a detailed discussion of how to perform linear regression with a large number of parameters see Ref.~\citenum{Hogg_2021}. Although there is a level of interpretability that can be gained from these techniques, it is not equivalent to that for simplistic linear models. These approaches should therefore be used when complex non-linearity is necessary for a solution.  

\section{Discussion}

Non-linear models have shown to be highly accurate predictive solutions for many problems. However, if equivalently accurate linear models can also be created this is often the preferable solution as they are directly interpretable. Depending on the application, one technique may be more advantageous. 

Alternatively, using both techniques together can have benefits. For instance, linear models can provide a baseline level of accuracy to compare to more complex techniques. An example is presented in Ref.~\citenum{Hansen2015}, where linear and polynomial models were created to show that there is significant improvement in the prediction of molecular properties when kernel ridge regression is used. These baselines may also incorporate knowledge about different transforms and interactions that are expected to occur.  Additionally, by analysing a non-linear model with interpretability techniques possible transforms can be exposed that can be used in a linear model.  

If a researcher wants to use non-linear ML models to improve predictive performance, taking a number of considerations into account can help to assess if improvements are likely to occur. This can be summarized as five key questions:  
\begin{itemize}
\item Do I expect non-linearity and interactions with my choice of features?
\item Are the majority of variables discrete with very few unique values?
\item For variables where non-linear relationships are expected, are these the key variables for the outcome and what is their distribution?
\item What is the size of the dataset and how does it compare to the number of variables used?
\end{itemize}

The flexible functional form of non-linear ML models has led to impressive improvements in predictive accuracy for many applications. However, not all problems require complex non-linear solutions. It is important to understand the reasons for this as it can prevent the unnecessary search for improvements in accuracy that will probably not be found. It can also prevent ``black box'' models being used for applications that have directly interpretable solutions. We have used simulated datasets and real datasets from across the sciences to explain when and why improvements with non-linear models may occur and when they are unlikely to not. In doing so, we have provided a unique discussion about the expected benefits of moving to non-linear models.

\section*{Acknowledgments}
The authors acknowledge funding from the IAS-Luxembourg (Audacity Grant DSEWELL). AA thanks the Center for Nonlinear Studies (CNLS) for resources and funding. Los Alamos National Laboratory LA-UR: LA-UR-22-30283.

\section*{Ethics declarations}
The authors declare no competing interests.

\section*{Data Availability Statement}
The data for predicting height is from the British Cohort Study (BCS)~\cite{Sullivan2022}. 
This dataset can be found through the UK Data Service at https://discover.ukdataservice.ac.uk/series/?sn=200001. 
The dataset for the siRNA example is taken from Ref.~\citenum{Huesken2005}. 
The dataset for the prediction of the formation energy is taken from Ref.~\citenum{Faber2015}.

\section*{Code Availability Statement}

The software required to make Fig.6, Fig.7 and Fig.10 is available at https://github.com/aa840/icepd.

\bibliography{refs}

\end{document}


\clearpage
\maketitle
\tableofcontents

\clearpage

\section{Supplementary Methods}

\subsection{Predicting Height}

\begin{figure*}[ht!]
    \centering
    \includegraphics[width=4.500in]{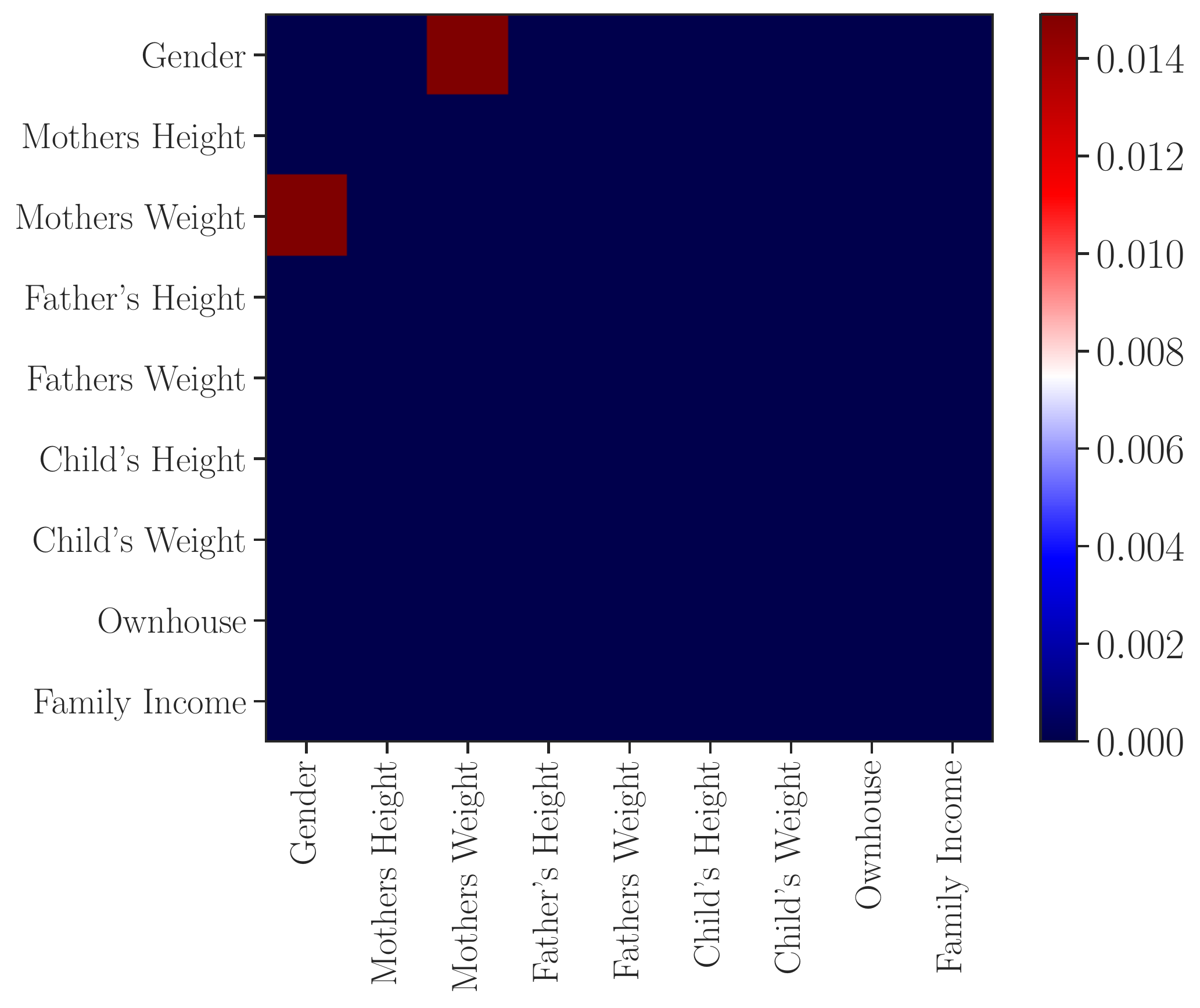}
    \caption{The interactions identified by the PD H-statistics for the neural network model for predicting height.   }
    \label{fig:Height_inter}
\end{figure*}

\begin{figure*}[ht!]
    \centering
    \includegraphics[width=2.000in]{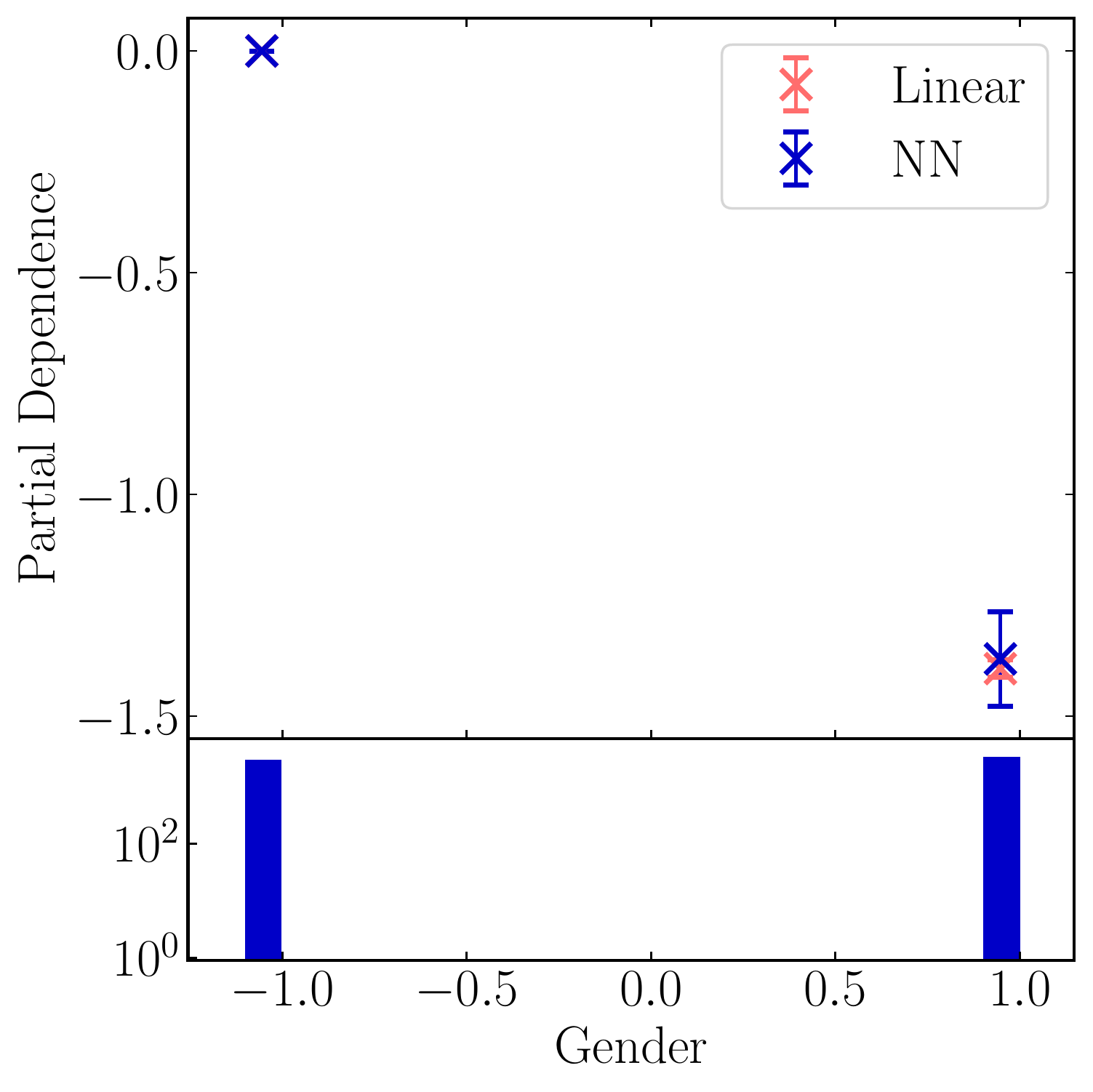}
    \includegraphics[width=2.000in]{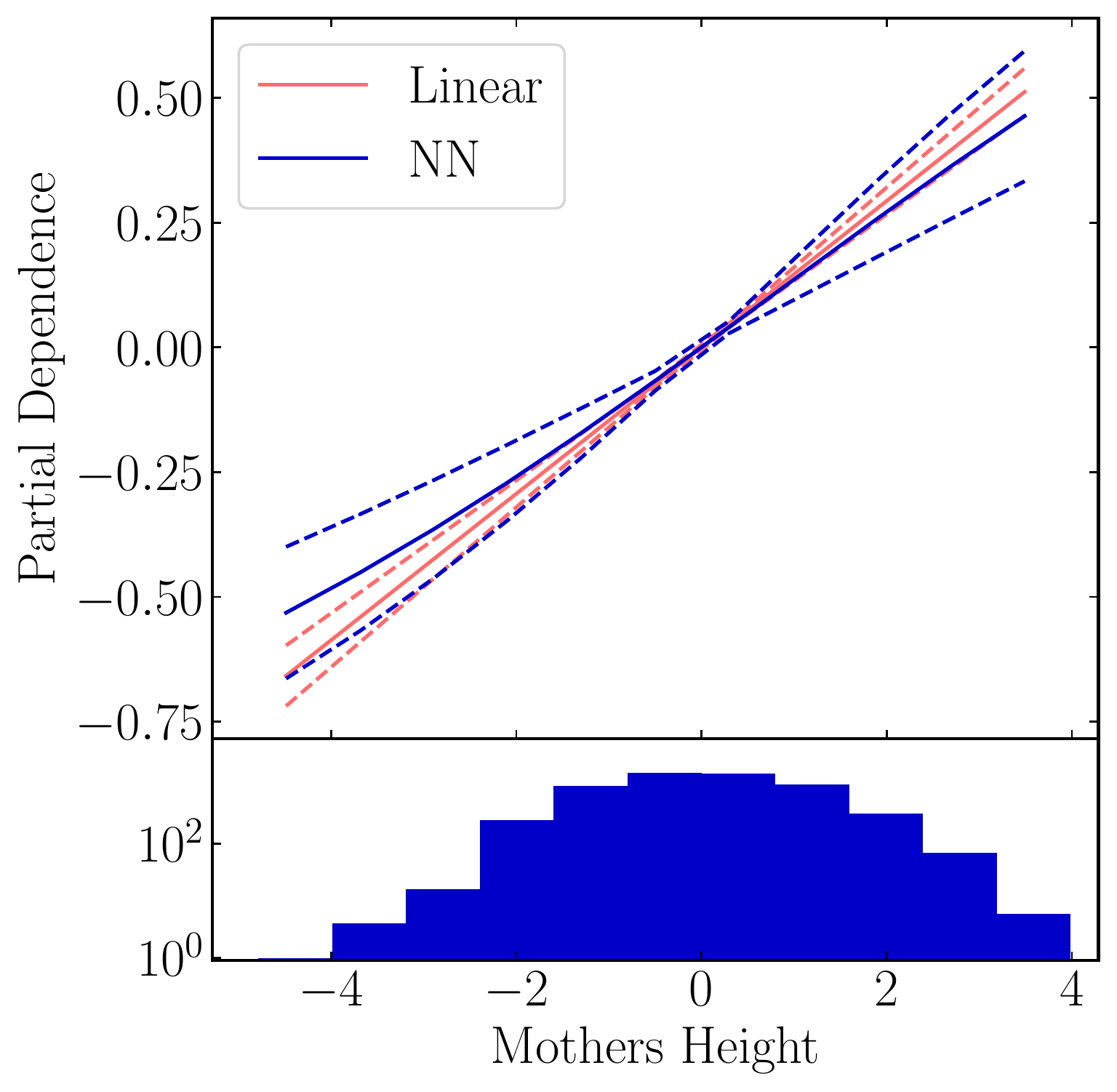}
    \includegraphics[width=2.000in]{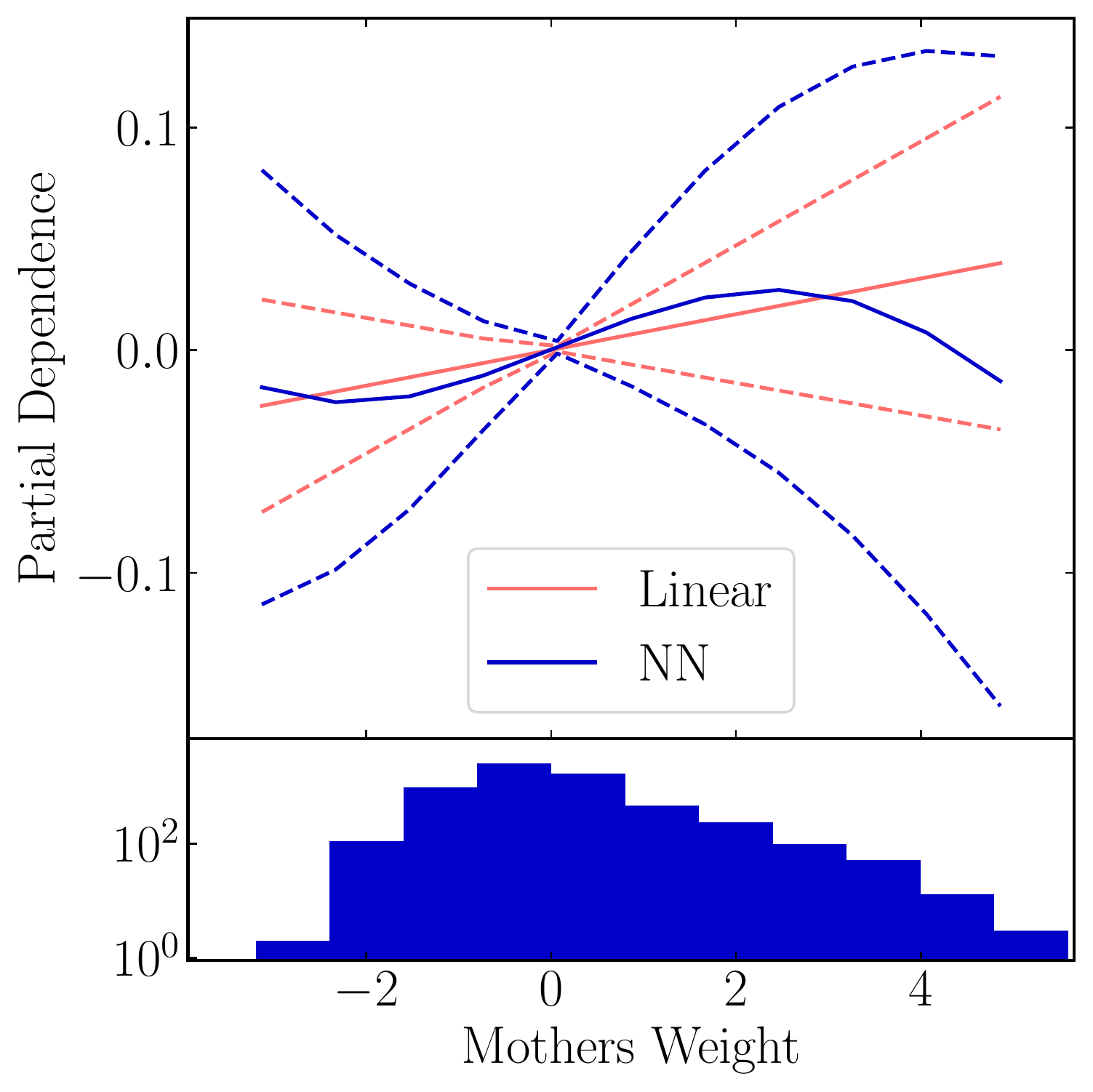}
    \includegraphics[width=2.000in]{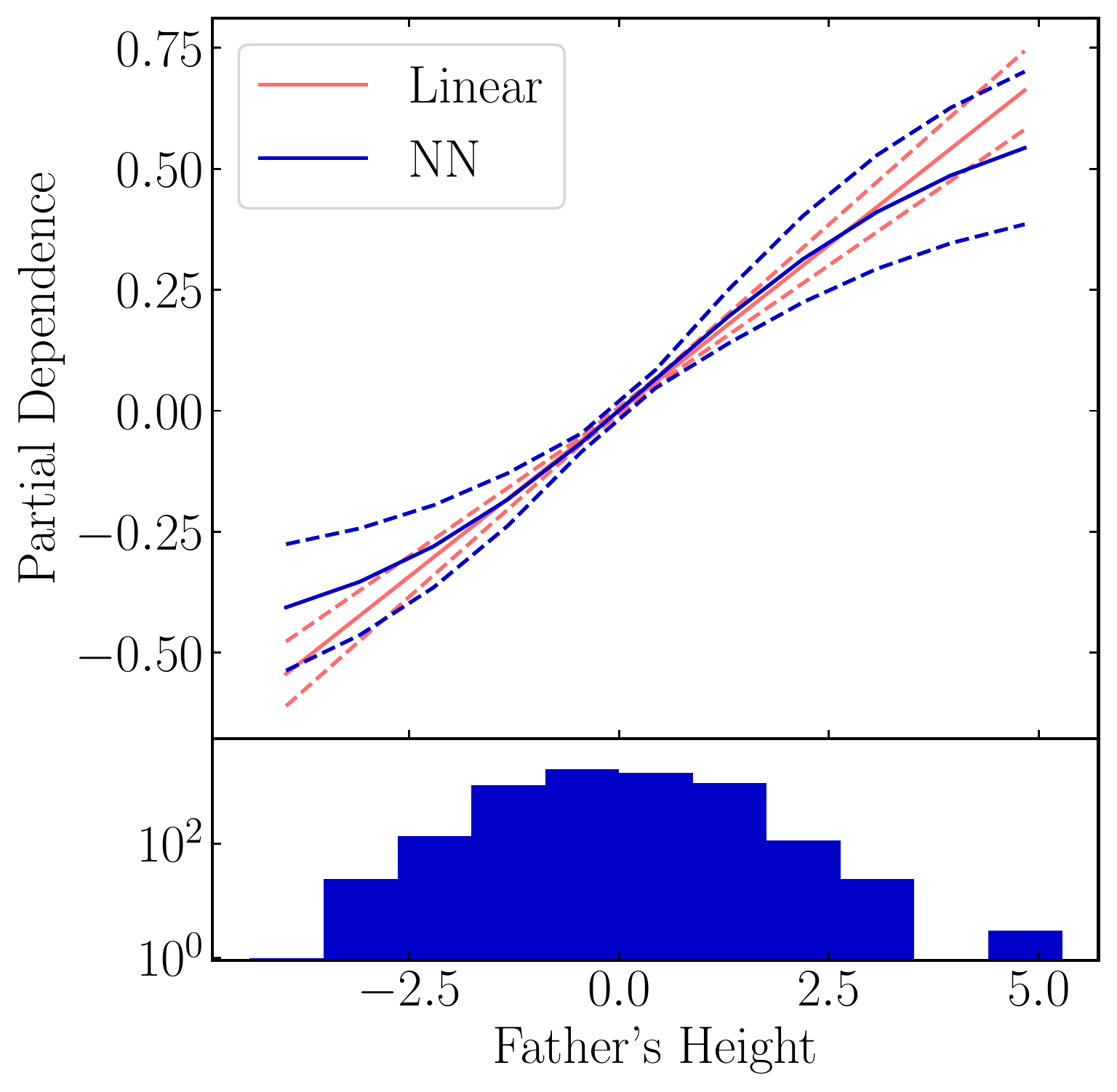}
    \includegraphics[width=2.000in]{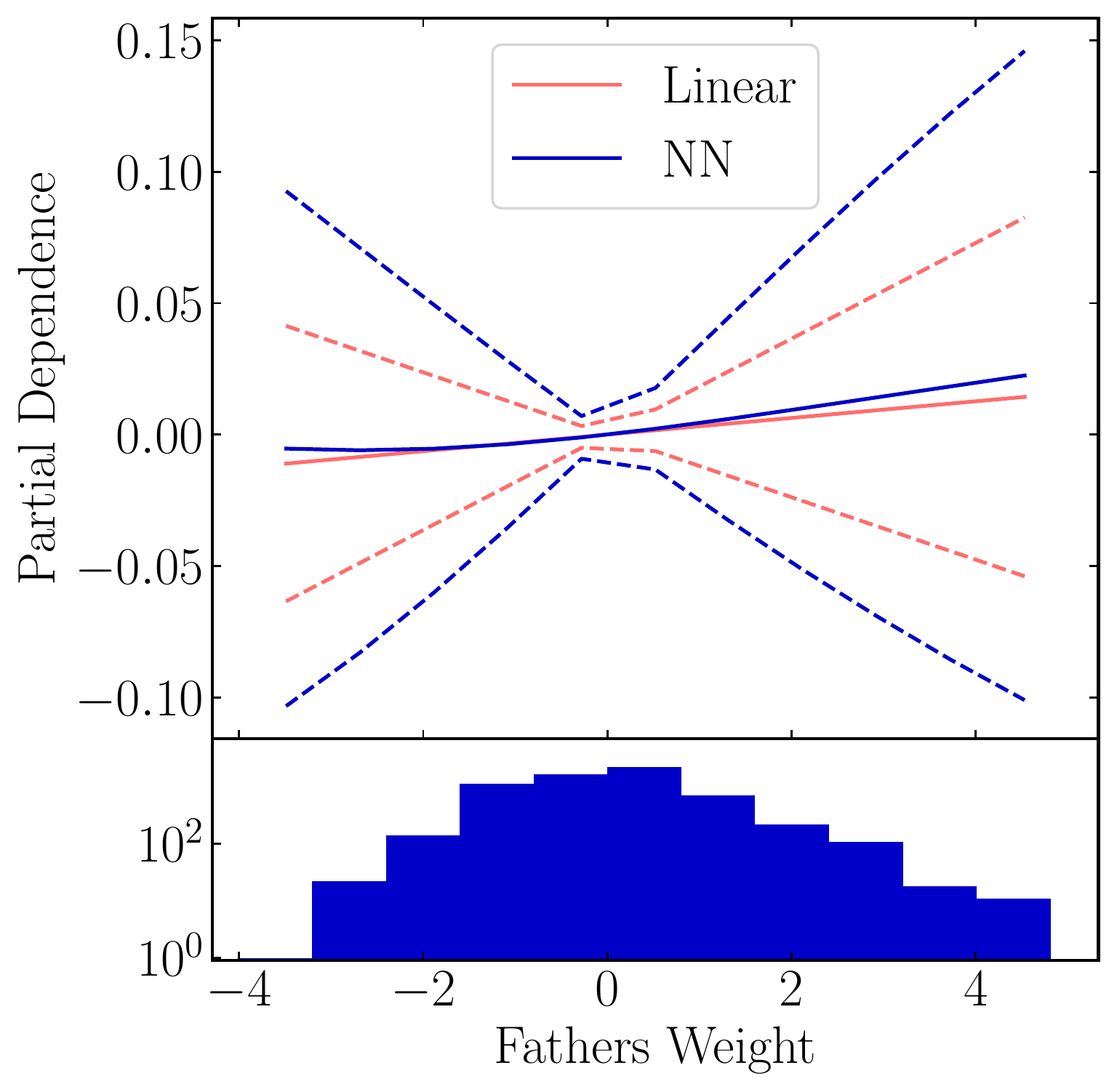}
    \includegraphics[width=2.000in]{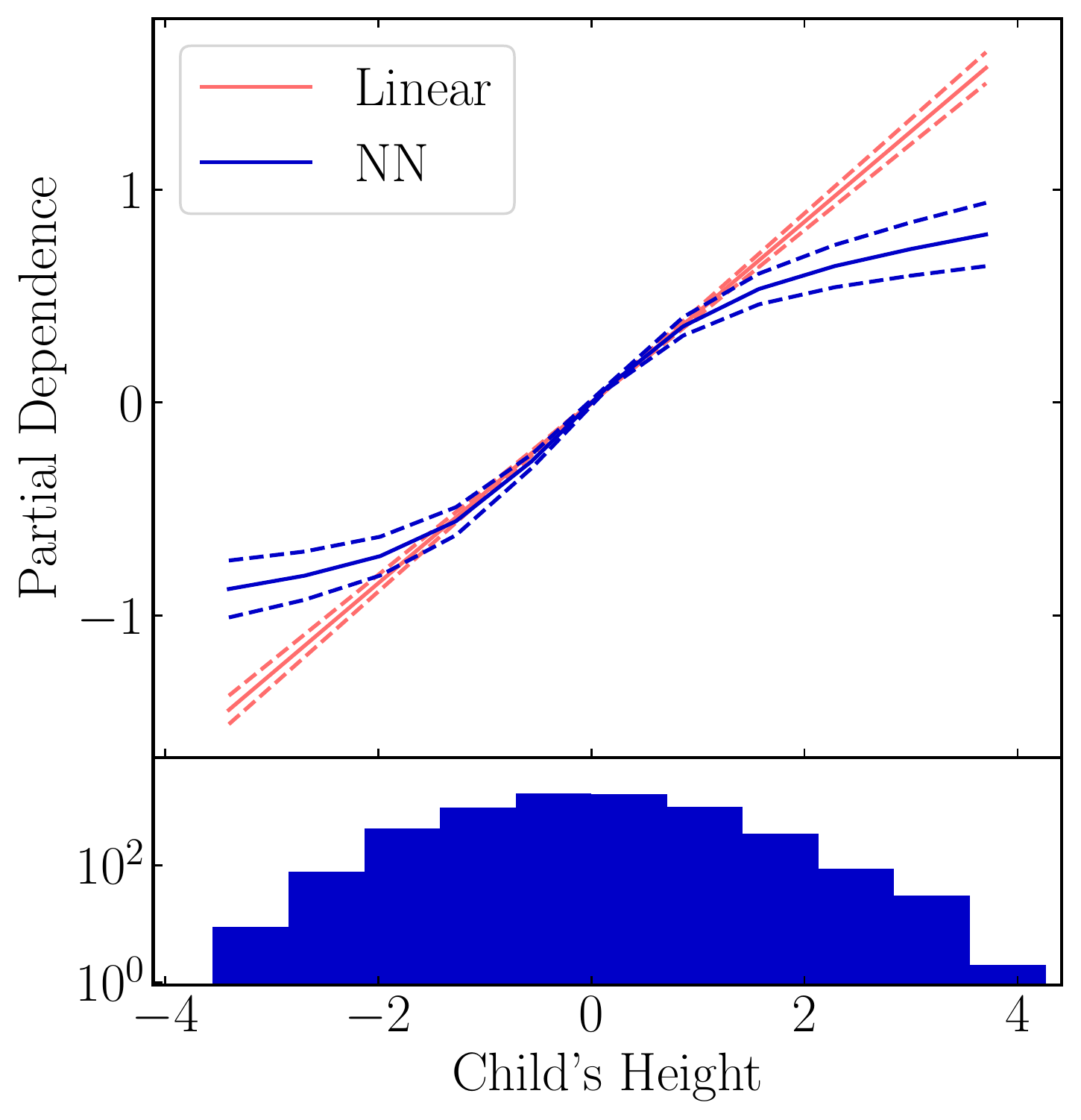}
    \includegraphics[width=2.000in]{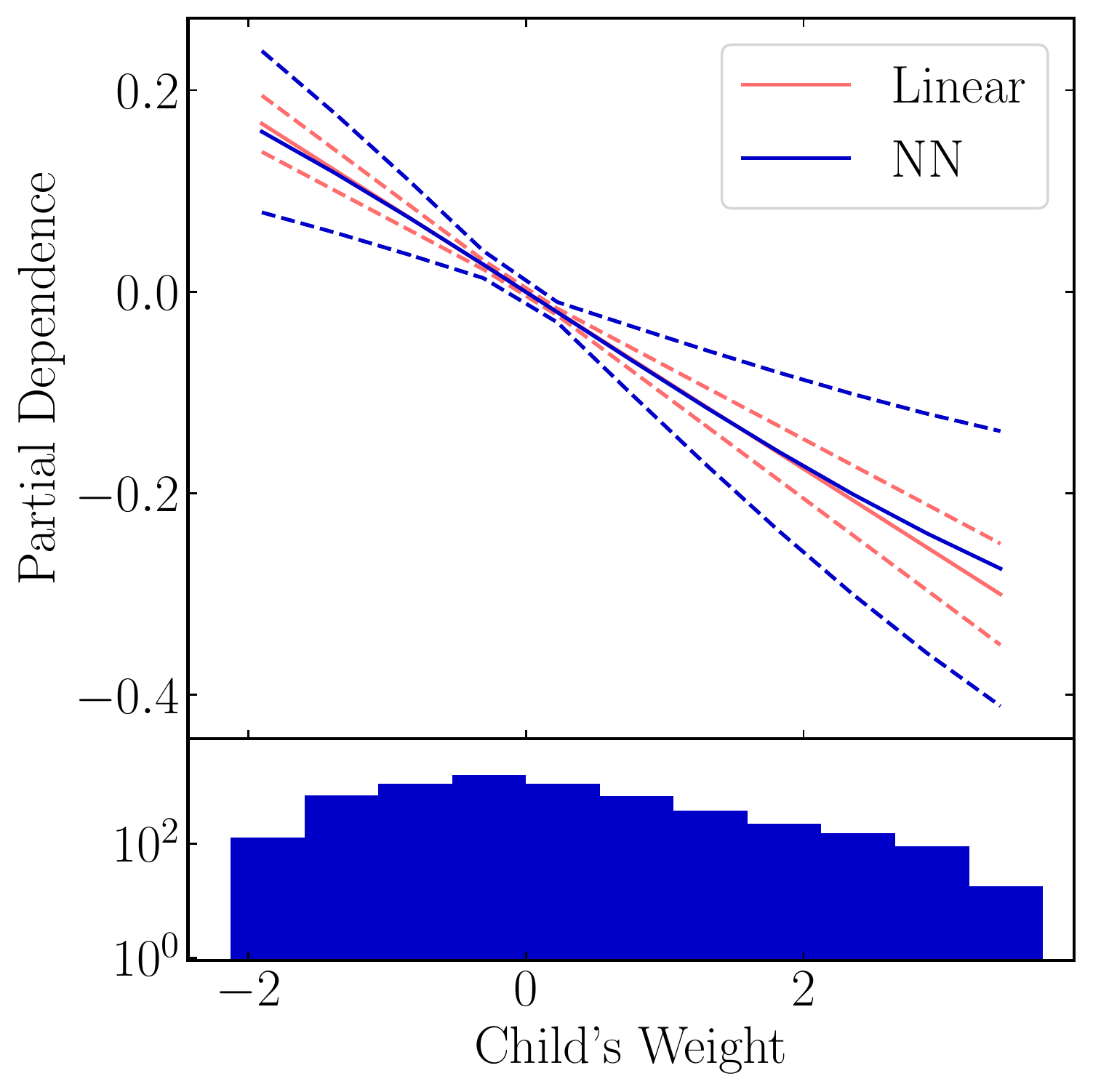}
    \includegraphics[width=2.000in]{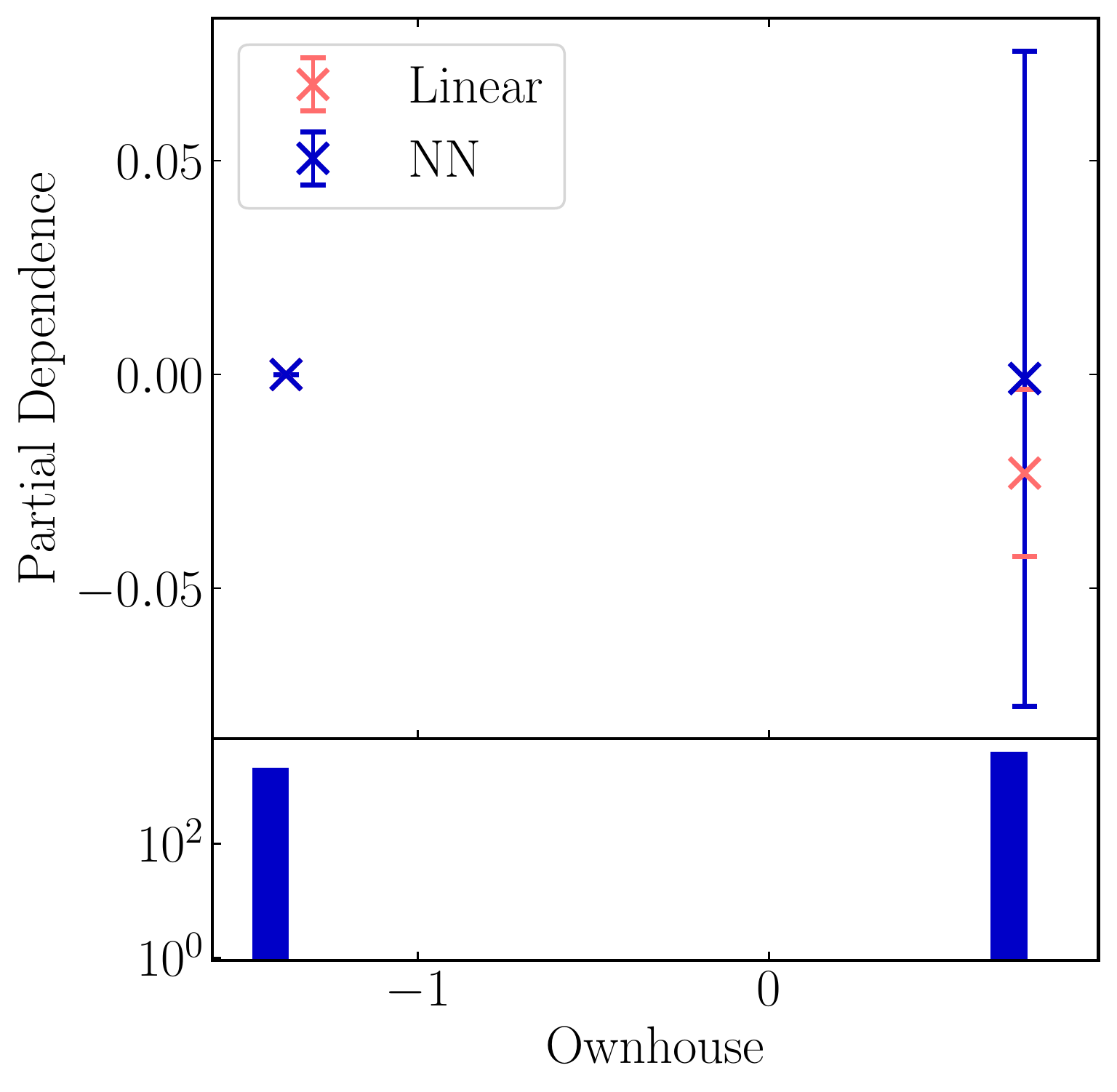}
    \includegraphics[width=2.000in]{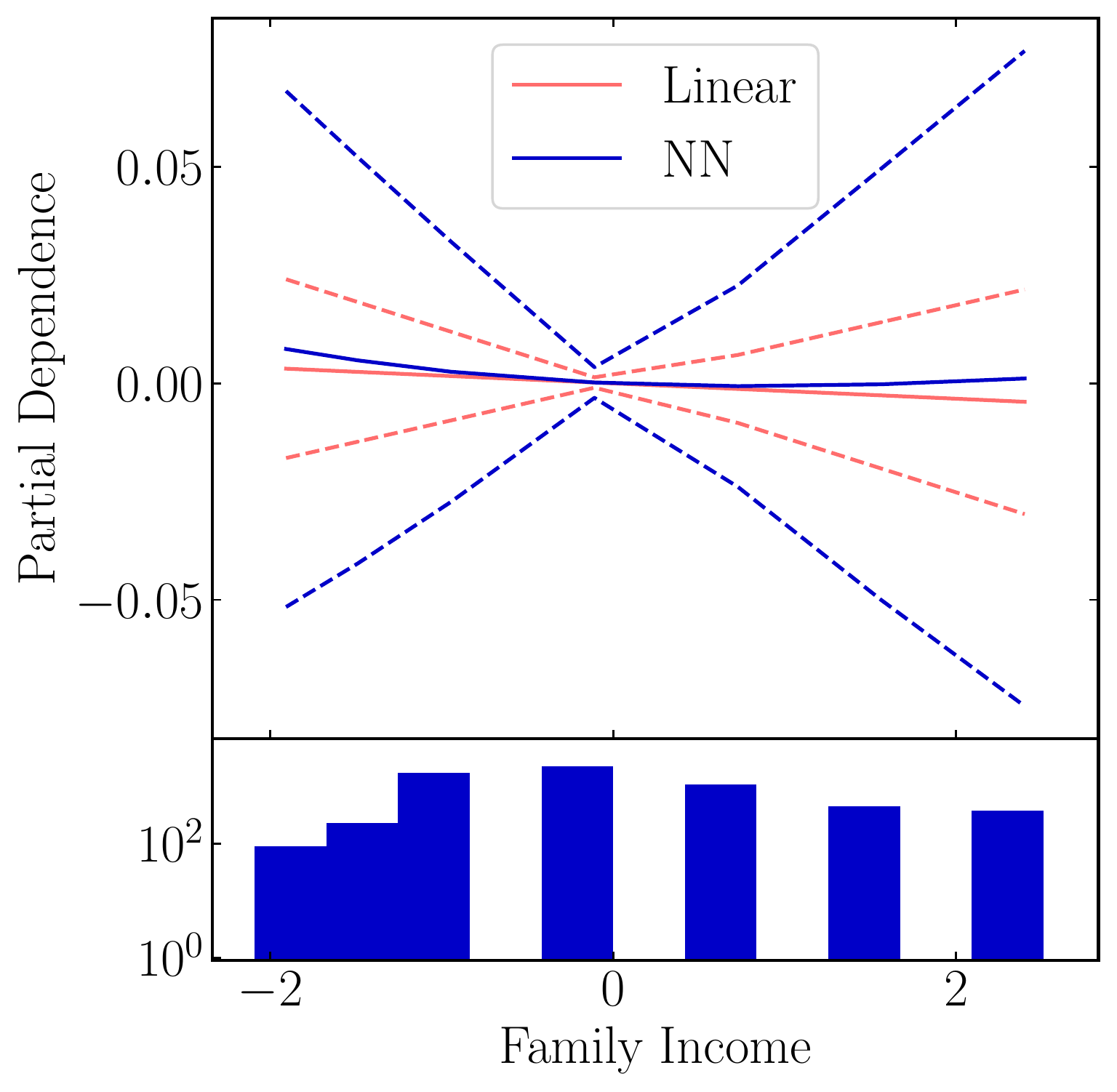}
    \caption{The PD plots for the neural network model for predicting height.   }
    \label{fig:Height_inter}
\end{figure*}

\begin{figure*}[ht!]
    \centering
    \includegraphics[width=5.000in]{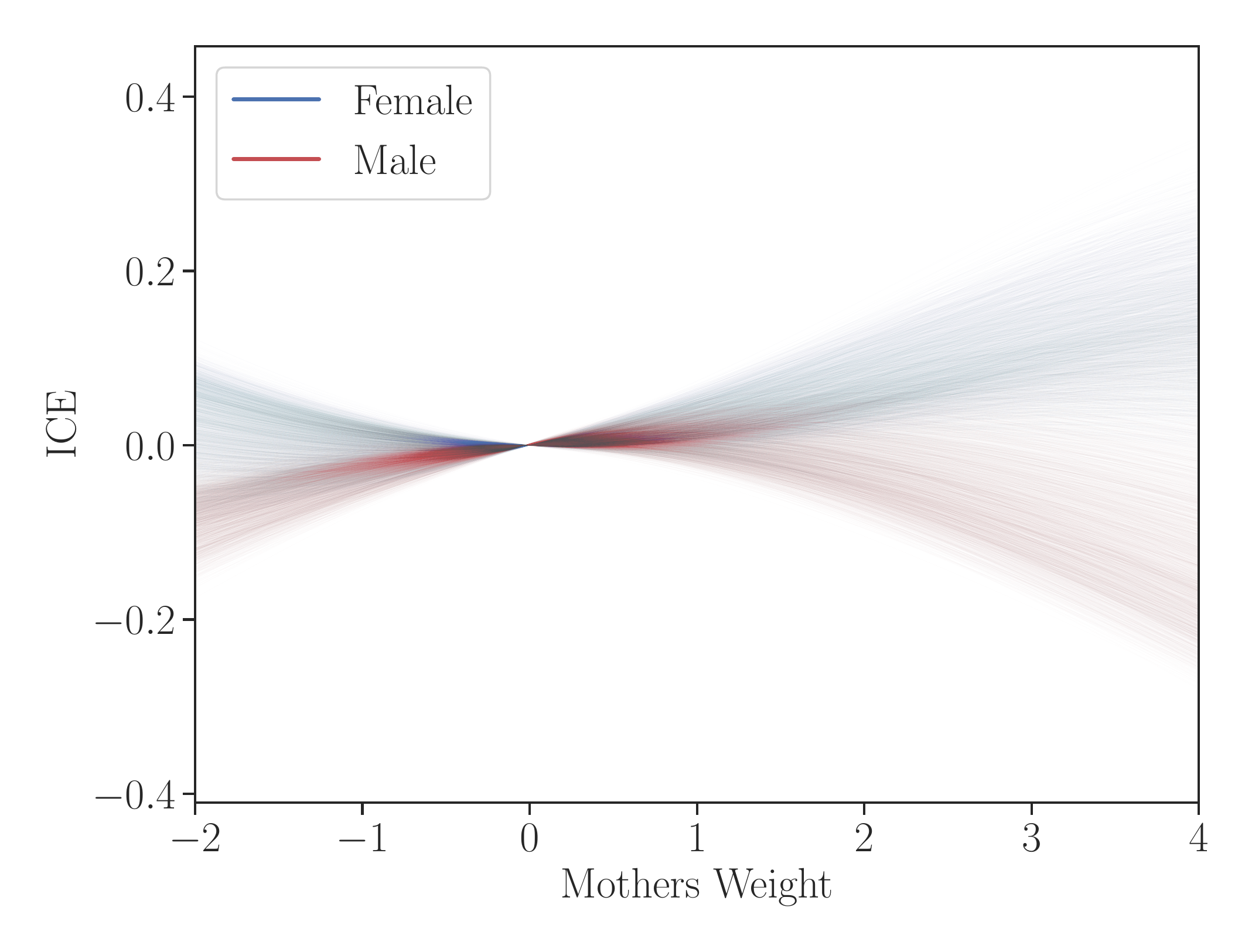}
    \caption{The ICE plot for the mother's weight for predicting the child's height at age 34 with the gender of the child shown by the color of the line.   }
    \label{fig:Height_inter}
\end{figure*}

\clearpage

\subsection{siRNA}

\subsubsection{Datasets}
A dataset of 2431 siRNA sequences is provided in Ref.~\citenum{Huesken2005} along with experimentally determined inhibitory activity. This dataset has been widely used for training and testing predictive models. The training/testing set split used in this work is the same as in Ref.~\citenum{Huesken2005}.

\subsubsection{Neural Network Model}
A feedforward NN, with one hidden layer was used for the simulated data and real datasets. A sigmoid activation function was employed and $l_1$ regularization was used on the NN weights and bias for the hidden layer only. The number of neurons in the hidden layer is set to 10 to recreate the NN used in Ref.~\citenum{Huesken2005}. 

\subsubsection{Features}
The sequence data is coded as follows:
$A = \{1 , 0, 0, 0\}$, $C = \{0, 1, 0, 0 \}$, $G = \{0, 0, 1, 0 \}$ and $U = \{0, 0, 0, 1\}$ 
This follows common practice. Therefore, for the sequence based model the 21 base sequence is represented by 84 binary features.
 
\subsubsection{One Hot Encoded Features}
The sequence data is coded as binary values. Calculating the PD and H-statistics must therefore take this into account. The partial dependence becomes:
\begin{equation}
PD(x_s) = \frac{1}{n} \sum^n_{i=1} [\hat{f}({x_s,  x_{s,OHE}=0, x^{(i)}_c})] 
\label{equ:partial_dep_ohe}
\end{equation}
where $x_{s,OHE}$ are the OHE variables in the same category as $x_s$. For example, to calculate $PD(A_0=1)$ the variables $C_0, G_0, U_0$ must all be set to 0.

\begin{figure*}[ht!]
    \centering
    \includegraphics[width=5.000in]{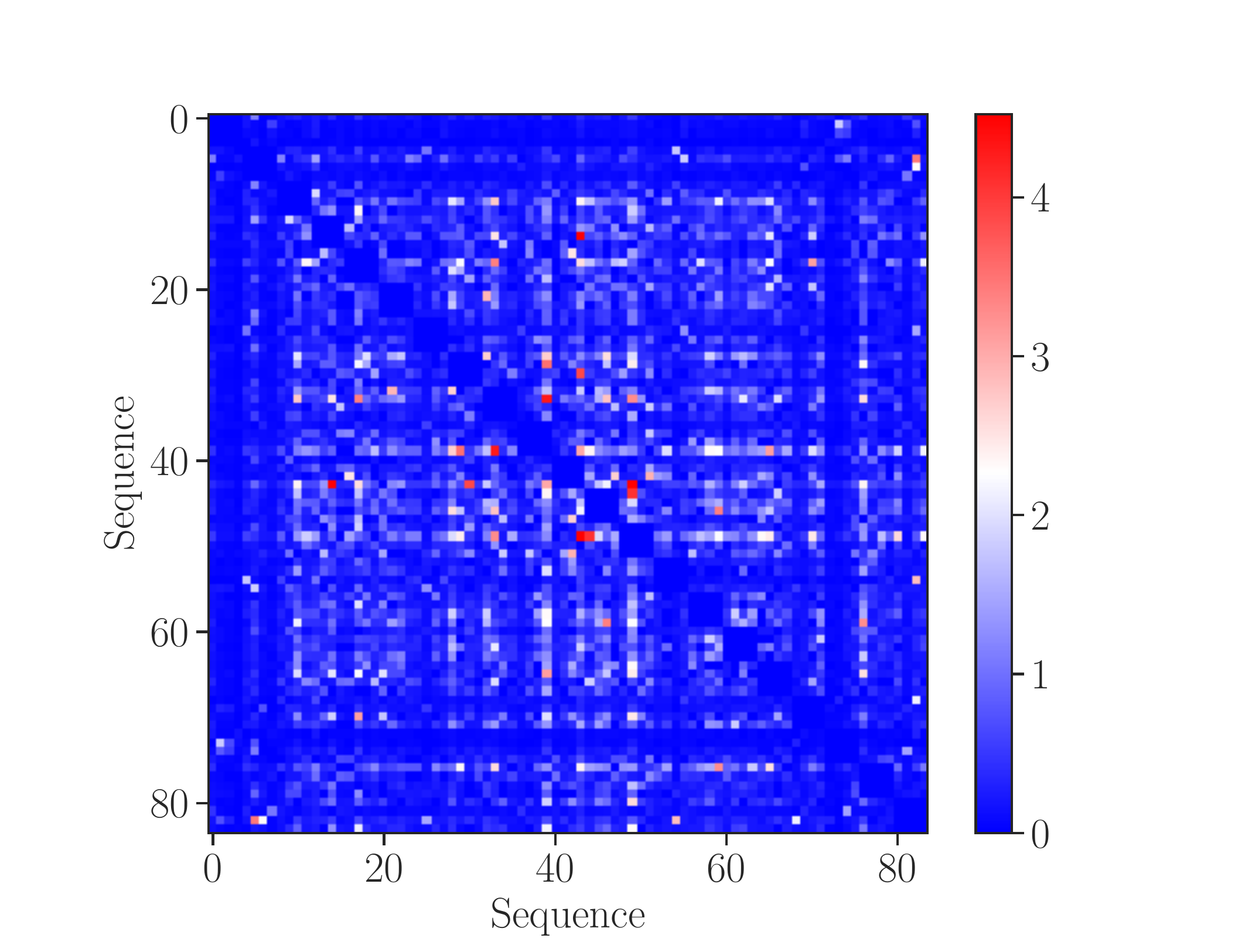}
    \caption{The interactions, computed by H-statistics, for the NN model for predicting the inhibitory activity of siRNA.}
    \label{fig:Height_inter}
\end{figure*}